\documentclass[10pt,twocolumn,letterpaper]{article}

\usepackage{cvpr}
\usepackage{times}
\usepackage{epsfig}
\usepackage{graphicx}
\usepackage{amsmath}
\usepackage{amssymb}
\usepackage{url}
\usepackage{booktabs}
\usepackage{cite}
\usepackage[normalem]{ulem}

\usepackage{multirow}

\def\ie{\emph{i.e}\onedot} 
\def\eg{\emph{e.g}\onedot} 
\def\para#1{\smallskip\noindent{\bf{#1}}}
\def\enum#1{\smallskip\noindent{\em{#1}}}

\cvprfinalcopy 

\def\cvprPaperID{2096}

\ifcvprfinal\fi
\begin{document}

\title{InLoc: Indoor Visual Localization with Dense Matching and View Synthesis}

\author{
Hajime Taira$^1$ \quad Masatoshi Okutomi$^1$ \quad Torsten Sattler$^2$ \quad Mircea Cimpoi$^3$ \\[1pt] 
Marc Pollefeys$^{2,4}$ \quad Josef Sivic$^{3,5}$ \quad Tomas Pajdla$^3$ \quad Akihiko Torii$^1$ \\[3pt]
$^1$Tokyo Institute of Technology \quad
$^2$Department of Computer Science, ETH Z\"{u}rich
\\[1pt]
$^3$CIIRC, CTU in Prague\thanks{CIIRC - Czech Institute of Informatics, Robotics, and Cybernetics, Czech Technical University in Prague.} \quad 
$^4$Microsoft, Redmond \quad
$^5$Inria\thanks{WILLOW project, Departement d'Informatique de l'\'Ecole Normale Sup\'erieure, ENS/INRIA/CNRS UMR 8548, PSL Research University.} 
}
\date{October 2017}
\maketitle
\ifcvprfinal\thispagestyle{empty}\fi
\begin{abstract}
\noindent
We seek to predict the 6 degree-of-freedom (6DoF) pose of a query photograph with respect to a large indoor 3D map. The contributions of this work are three-fold. First, we develop a new large-scale visual localization method targeted for indoor environments. The method proceeds along three steps: (i) efficient retrieval of candidate poses that ensures scalability to large-scale environments, (ii) pose estimation using dense matching rather than local features to deal with textureless indoor scenes, and  (iii) pose verification by virtual view synthesis to cope with significant changes in viewpoint, scene layout, and occluders. Second, we collect a new dataset with reference 6DoF poses for large-scale indoor localization. Query photographs are captured by mobile phones at a different time than the reference 3D map, thus presenting a realistic indoor localization scenario. Third, we demonstrate that our method significantly outperforms current state-of-the-art indoor localization approaches on this new challenging data.
\end{abstract}

\section{Introduction \label{sec:intro}}
\begin{figure}[t]
    \centering
    {\footnotesize
    \includegraphics[width=0.99\linewidth]{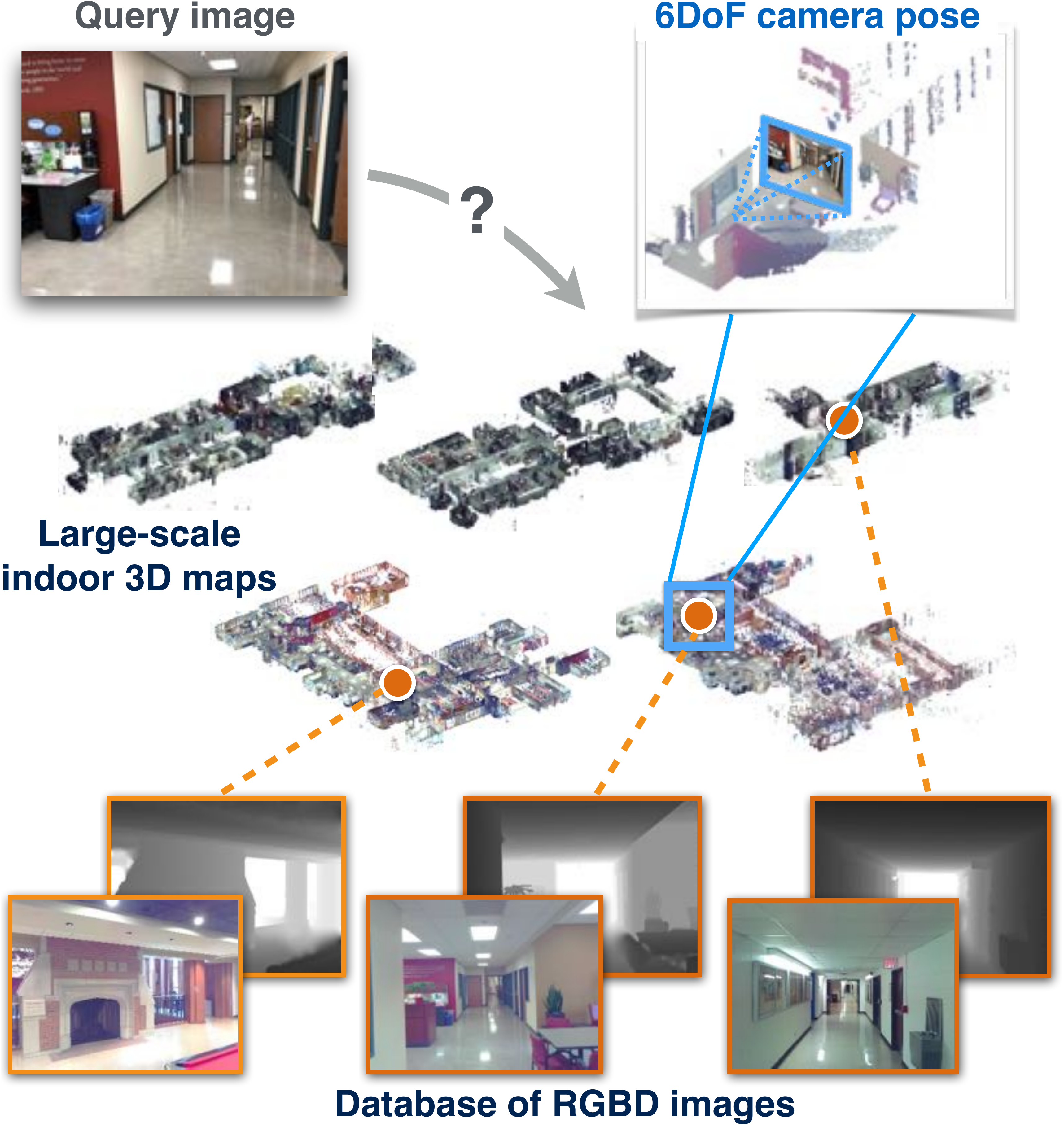}
    \caption{{\bf Large-scale indoor visual localization.} Given a database of geometrically-registered RGBD images, we predict the 6DoF camera pose of a query RGB image by retrieving candidate images, estimating candicate camera poses, and selecting the best matching camera pose. 
    To address inherent difficulties in indoor visual localization, we introduce the ``InLoc'' approach that performs a sequence of progressively stricter verification steps. 
    }
    }
    \label{fig:teaser}
\end{figure}

\noindent Autonomous navigation inside buildings is a key ability of robotic intelligent systems~\cite{debski2015open,lim2012real}. 
Successful navigation requires both to localize a robot and to determine a path to its goal. 
One approach to solving the localization problem is to build a 3D map of the building and then use a camera\footnote{While RGBD sensors could also be used indoors, they are often too energy-consuming for mobile scenarios or have only a short-range to scan close-by objects (faces). Thus, purely RGB-based localization approaches are also relevant in indoor scenes. Obviously, indoor scenes are GPS-denied environments.} 
to estimate the current position and orientation of the robot (Figure~\ref{fig:teaser}). Imagine also the benefit of an intelligent indoor navigation system that helps you find your way, for example, at Chicago airport, Tokyo Metropolitan station or the CVPR conference center.
Besides intelligent systems, the \emph{visual localization} problem is also highly relevant for any type of Mixed Reality application, including Augmented Reality~\cite{middelberg2014scalable,wagner2010real,Castle08ISWC}. 

Due to the availability of datasets, \eg, obtained from Flickr~\cite{li2012worldwide} or captured from autonomous vehicles~\cite{chen2011city,RobotCarDatasetIJRR}, large-scale localization in urban environments has been an active field of research~\cite{sattler2018benchmark,sattler2017efficient,li2012worldwide,irschara2009structure,camposeco2017toroidal,cao2014minimal,kendall2017geometric,Zeisl2015ICCV,Svarm17PAMI,arandjelovic2014dislocation,Arandjelovic16,torii201524,chen2011city,choudhary2012visibility,Gronat13,middelberg2014scalable,sattler2015hyperpoints,sattler2016large,sattler2017large,torii2013visual,weyand2016planet,zamir2010accurate}. 
In contrast, indoor localization~\cite{shotton2013scene,lim2012real,Brachmann2017DSAC,Brachmann2016CVPR,valentin2016learning,sun2017dataset,Schmidt2017RAL,wang2015lost} has received less attention in the last years. 
At the same time, indoor localization is, in many ways, a harder problem than urban localization: 
1)~Due to the short distance to the scene geometry, even small changes in viewpoint lead to large changes in image appearance. 
For the same reason, ocluders such as humans or chairs often have a stronger impact compared to urban scenes. 
Thus, indoor localization approaches have to handle significantly larger changes in appearance between a query and reference images. 
2)~Large parts of indoor scenes are textureless and textured areas are typically rather small. 
As a result, feature matches are often clustered in small regions of the images, resulting in unstable pose estimates~\cite{irschara2009structure}. 
3)~To make matters worse, buildings are often highly symmetric with many repetitive elements, both on large (similar corridors, rooms, \etc) and small (similar chairs, tables, doors \etc) scale. 
While structural ambiguities also cause problems in urban environments, they often only occur in larger scenes~\cite{torii2013visual,arandjelovic2014dislocation,sattler2016large}. 
4) The appearance of indoor scenes changes considerably over the course of a day due to the complex illumination conditions (indirect light through windows and active illumination from lamps). 
5) Indoor scenes are often highly dynamic over time as furniture and personal effects are moved through the environment. 
In contrast, the overall appearance of building facades does not change too much over time. 

This paper addresses these difficulties inherent to indoor visual localization by proposing a new localization method. 
Our approach starts with an image retrieval step, using a compact image representation~\cite{Arandjelovic16} that scales to large scenes. 
Given a shortlist of potentially relevant database images, we apply two progressively more discriminative geometric verification steps:
(i) We use dense matching of CNN descriptors that capture spatial configurations of higher-level structures (rather than individual local features) to obtain the correspondences required for camera pose estimation. 
(ii) We then apply a novel pose verification step based on virtual view synthesis that can
accurately verify whether the query image depicts the same place by dense pixel-level matching, again not relying on sparse local features.

Historically, the datasets used to evaluate indoor visual localization were restricted to small, often room-scale, scenes. 
Driven by the interest in semantic scene understanding~\cite{Dai2017CVPR,armeni_cvpr16,xiao2013sun3d} and enabled by scalable reconstruction techniques~\cite{niessner2013hashing,newcombe2011kinectfusion,Halber2017CVPR}, large-scale indoor datasets covering multiple rooms or even whole buildings are becoming available~\cite{Chang20173DV,Dai2017CVPR,wijmans17rgbd,sun2017dataset,xiao2013sun3d,armeni_cvpr16,xiao2014reconstructing,wang2015lost}. However, most of these datasets focus on reconstruction~\cite{wijmans17rgbd,xiao2014reconstructing} and semantic scene understanding~\cite{Dai2017CVPR,Chang20173DV,xiao2013sun3d,armeni_cvpr16} and are not suitable for localization. To address this issue, we create a new dataset for indoor localization that, in contrast to other existing indoor localization datasets~\cite{sun2017dataset,glocker2013real,armeni_cvpr16}, has two important properties. First, the dataset is large-scale, capturing two university buildings. Second, the query images are acquired using a smartphone at a time months apart from the date of capture of the reference 3D model. As a result, the query images and the reference 3D model often contain large changes in scene appearance due to the different layout of furniture, occluders (people), and illumination, representing a realistic and challenging indoor localization scenario.

\para{Contributions.}
Our contributions are three-fold. First, we develop a novel visual localization approach suitable for large-scale indoor environments. The key novelty of our approach lies in carefully introducing dense feature extraction and matching in a sequence of progressively stricter verification steps.       
To the best of our knowledge, the present work is the first to clearly demonstrate the benefit of dense data association for indoor localization. 
Second, we create a new dataset suitably designed for large-scale indoor localization that contains large variation in appearance between queries and the 3D database due to large viewpoint changes, moving furniture, occluders or changing illumination. The query images are taken at a different time from the reference database, using a handheld device, and at different moments of the day, to capture enough variability, bridging the gap to realistic usage scenarios. 
The code and data are publicly available on the project page~\cite{projectpage}. 
Third, the proposed method shows a solid improvement over existing state-of-the-art results, showing an \textbf{absolute improvement of 17--20\%} in the percent of correctly localized queries  within a 0.25 -- 0.5 m error, which is of high importance for indoor localization.

\section{Related work \label{sec:related}}
\noindent We next review previous work on visual localization.

\para{Image retrieval based localization.}
Visual localization in large-scale urban environments is often approached as an image retrieval problem. 
The location of a given query image is predicted by transferring the geotag of the most similar image retrieved from a geotagged database~\cite{chen2011residual,arandjelovic2014dislocation,torii2013visual,torii201524,sattler2016large,Arandjelovic16,kim2017learned}. This approach scales to entire cities thanks to compact image descriptors and efficient indexing techniques~\cite{arandjelovic2013all,arandjelovic2012three,chum2007total,jegou2008hamming,jegou2012aggregating,nister2006scalable,sivic2003video,van2010visual} and can be further improved by spatial re-ranking~\cite{philbin2007object},  informative feature selection~\cite{chum2007total,chum2011total} or feature weighting~\cite{jegou2009burstiness,sattler2016large,torii2013visual,Gronat13}. Most of the above methods are based on image representations using sparsely sampled local invariant features. While these representations have been very successful, outdoor image-based localization has recently also been approached  using {\em densely sampled} local descriptors~\cite{torii201524} or (densely extracted) descriptors based on convolutional neural networks~\cite{Arandjelovic16,kim2017learned,lin2015bilinear,weyand2016planet}. 
However, the main shortcoming of all the above methods is that they output only an approximate location of the query, not an exact 6DoF pose.

\para{Visual localization using 3D maps. }
Another approach is to directly obtain 6DoF camera pose with respect to a pre-built 3D map. The map is usually composed of a 3D point cloud constructed via Structure-from-Motion (SfM)~\cite{agarwal2011building} where each 3D point is associated with one or more local feature descriptors. The query pose is then obtained by feature matching and solving a Perspective-n-Point problem (PnP)~\cite{sattler2017efficient,li2012worldwide,choudhary2012visibility,irschara2009structure,camposeco2017toroidal,cao2014minimal,sattler2015hyperpoints,kendall2017geometric}. 
Alternatively, pose estimation can be formulated as a learning problem, where the goal is to train a regressor from the input RGB(D) space to camera pose parameters~\cite{Brachmann2017DSAC,shotton2013scene,kendall2017geometric,Walch2017ICCV}. While promising, scaling these methods to large-scale datasets is still an open challenge.

\para{Indoor 3D maps. }
Indoor scene datasets~\cite{quattoni2009recognizing,Singh2012DiscPat,pandey2011scene,torralba2003context} have been introduced for tasks such scene recognition, classification, and object retrieval. 
With the increased availability of laser range scanners and time-of-flight (ToF) sensors, several datasets include depth data besides RGB images~\cite{lai2014unsupervised,anand2013contextually,silberman2012indoor,glocker2013real,xiao2013sun3d,armeni_cvpr16,Dai2017CVPR} and some of these datasets also provide reference camera poses registered into the 3D point cloud~\cite{glocker2013real,xiao2013sun3d,armeni_cvpr16}, though their focus is not on localization. Datasets focused specifically on indoor localization~\cite{shotton2013scene,valentin2016learning,sun2017dataset} have so far captured fairly small spaces such as a single room (or a single floor at largest) and have been constructed from densely-captured sequences of RGBD images. 
More recent datasets~\cite{wijmans17rgbd,Chang20173DV} provide larger scale (multi-floor) indoor 3D maps containing RGBD images registered to a global floor map. However, they are designed for object retrieval, 3D reconstruction, or training deep-learning architectures. 
Most importantly, they do not contain query images taken from viewpoints far from database images, which are necessary for evaluating visual localization.

To address the shortcomings of the above datasets for large-scale indoor visual localization, we introduce a new dataset that includes query images captured at a different time from the database, taken from a wide range of viewpoints, with a considerably larger 3D database distributed across multiple floors of multiple buildings. Furthermore, our dataset contains various difficult situations for visual localization, \eg, textureless and highly symmetric office scenes, repetitive tiles, and repetitive objects that confuse the existing visual localization methods designed for outdoor scenes. The newly collected dataset is described next. 

\section{The InLoc dataset for visual localization \label{sec:dataset}}
\noindent Our dataset is composed of a database of RGBD images geometrically registered to the floor maps augmented with a separate set of RGB query images taken by hand-held devices to make it suitable for the task of indoor localization (Figure~\ref{fig:datasetimgsamples}). The provided query images are annotated with manually verified ground-truth 6DoF camera poses (reference poses) in the global coordinate system of the 3D map. 

\para{Database.}
The base indoor RGBD dataset~\cite{wijmans17rgbd} consists of 277 RGBD panoramic images obtained from scanning two buildings at the Washington University in St.~Louis with a Faro 3D scanner. Each RGBD panorama has about 40M 3D points in color. The base images are divided into five scenes: DUC1, DUC2, CSE3, CSE4, and CSE5, representing five floors of the mentioned buildings, and are geometrically registered to a known floor plan~\cite{wijmans17rgbd}. 
The scenes are scanned sparsely on purpose, to cover a larger area with a small number of scans to reduce the required manual work, as well as due to the long operating times of the high-end scanner used. The area per scan varies between 23.5 and 185.8 $m^2$. 
This inherently leads to critical view changes between query and database images when compared with other existing datasets~\cite{wang2015lost,sun2017dataset,valentin2016learning}\footnote{
For example, in the database of~\cite{sun2017dataset}, the scans are distributed on one single floor, and the area per each database image is less than 45 $m^2$. }.

For creating an image database suitable for indoor visual localization evaluation, a set of perspective images is generated by following the best practices from outdoor visual localization~\cite{chen2011city,zamir2010accurate,torii201524}. We obtain 36 perspective RGBD images from each panorama by extracting standard perspective views ($60^\circ$ FoV) with a sampling stride of $30^\circ$ in yaw and $\pm30^\circ$ in pitch directions, resulting in 10K perspective images in total (Table~\ref{tab:cutoutdetail}).
Our database contains significant challenges, such as repetitive patterns (stairs, pillars), frequently appearing building structures (doors, windows), furniture changing position, people moving across the scene, 
and textureless and highly symmetric areas (walls, floors, corridors, classrooms, open spaces). 
\begin{table}[t]
    \centering
    {\footnotesize
    \begin{tabular}{r|ccc}
    & Number & Image size [pixel] & FoV [degree] \\[1pt] \hline
    Query & 356 & 4,032$\times$3,024 & 65.57 \\[1pt]
    Database & 9,972 & 1,600$\times$1,200 & 60 \\[3pt]
    \end{tabular}
    \caption{Statistics of the {\bf InLoc dataset}.}
    \label{tab:cutoutdetail}
    }
\end{table}
{\tabcolsep = 1pt
\begin{figure}[t]
    \centering
    {\footnotesize
    \begin{tabular}{cccc}
    \includegraphics[width=0.242\linewidth]{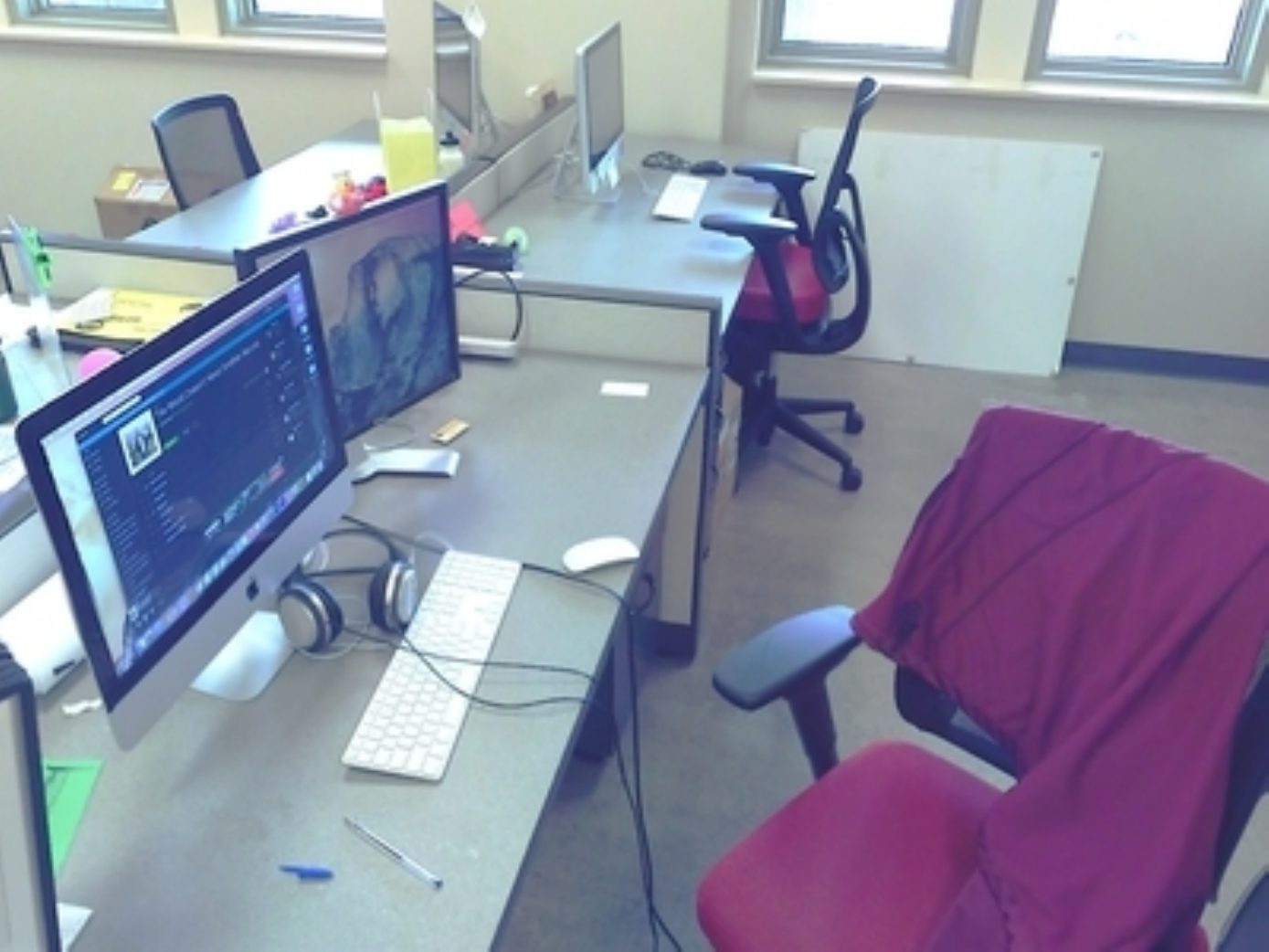} & 
    \includegraphics[width=0.242\linewidth]{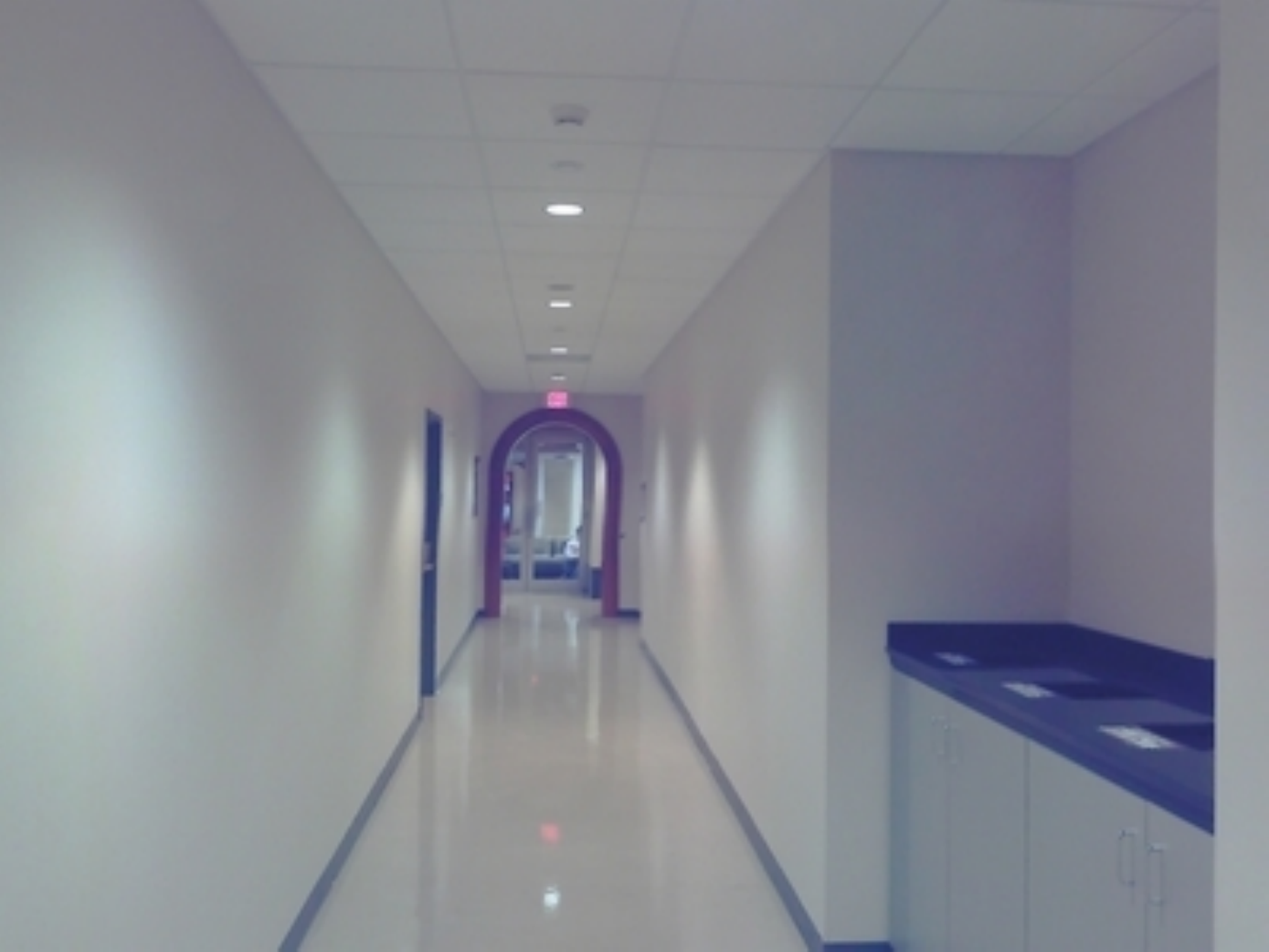} & 
    \includegraphics[width=0.242\linewidth]{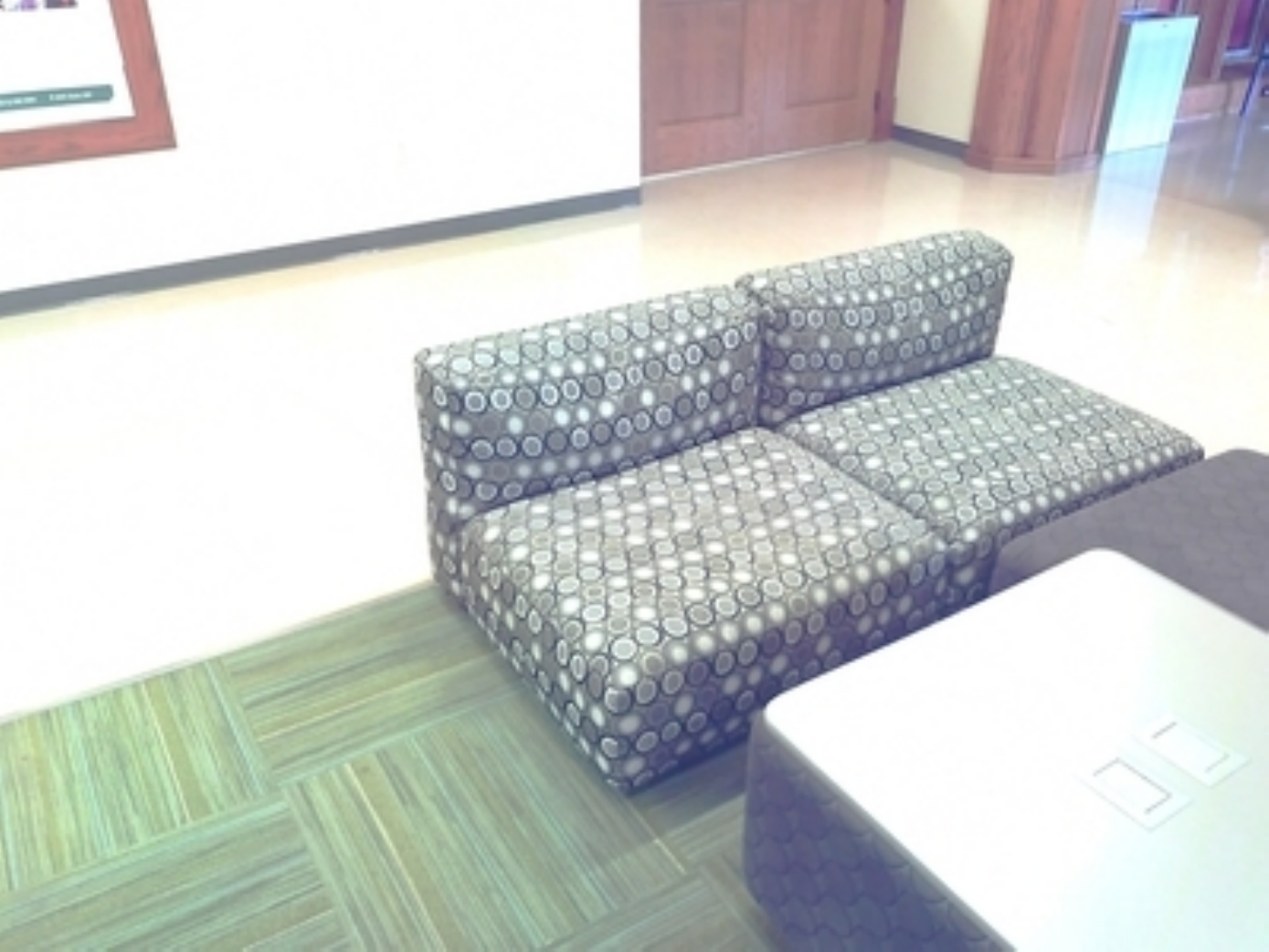} & 
    \includegraphics[width=0.242\linewidth]{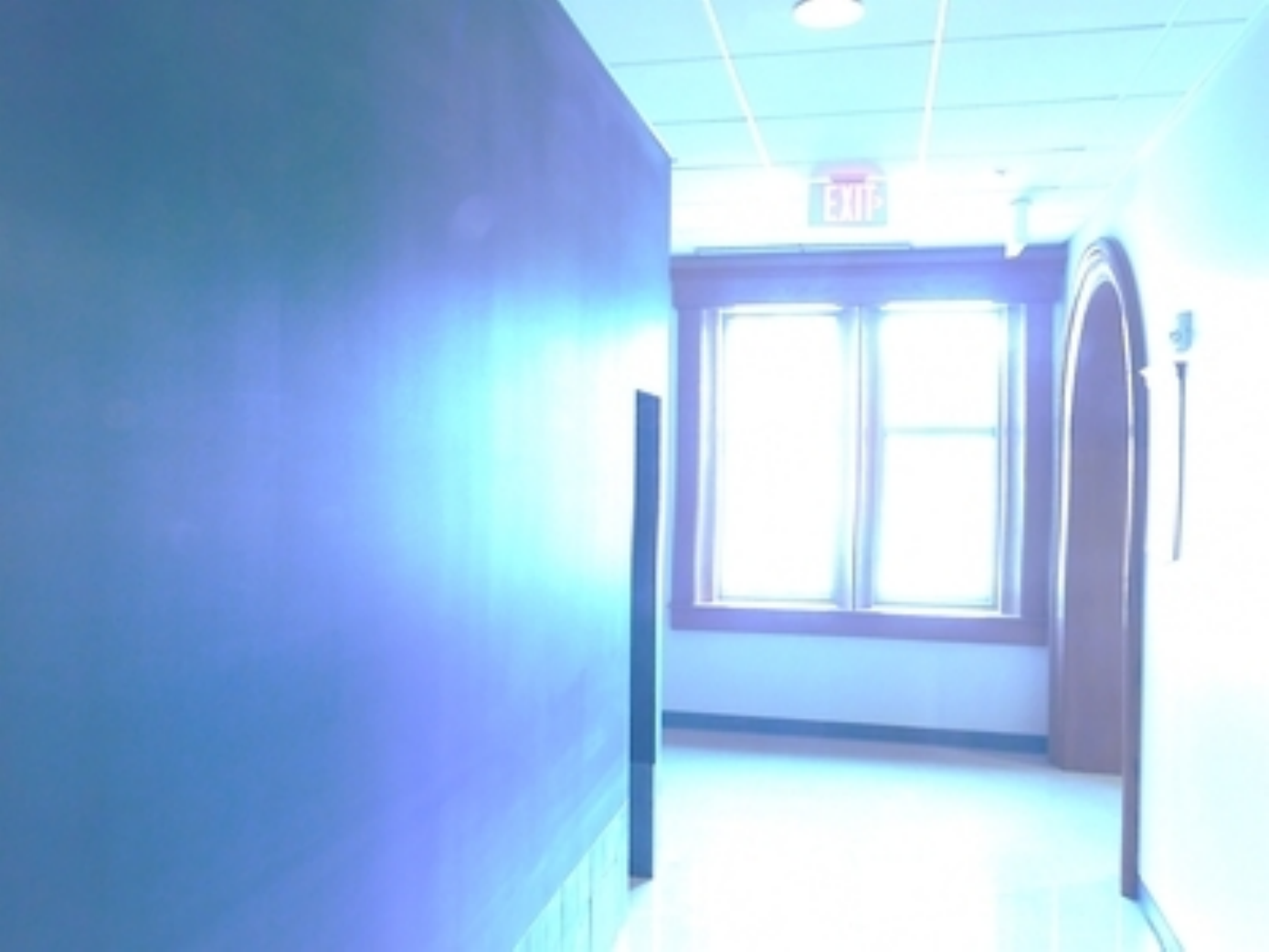} \\[-1pt]
    \includegraphics[width=0.242\linewidth]{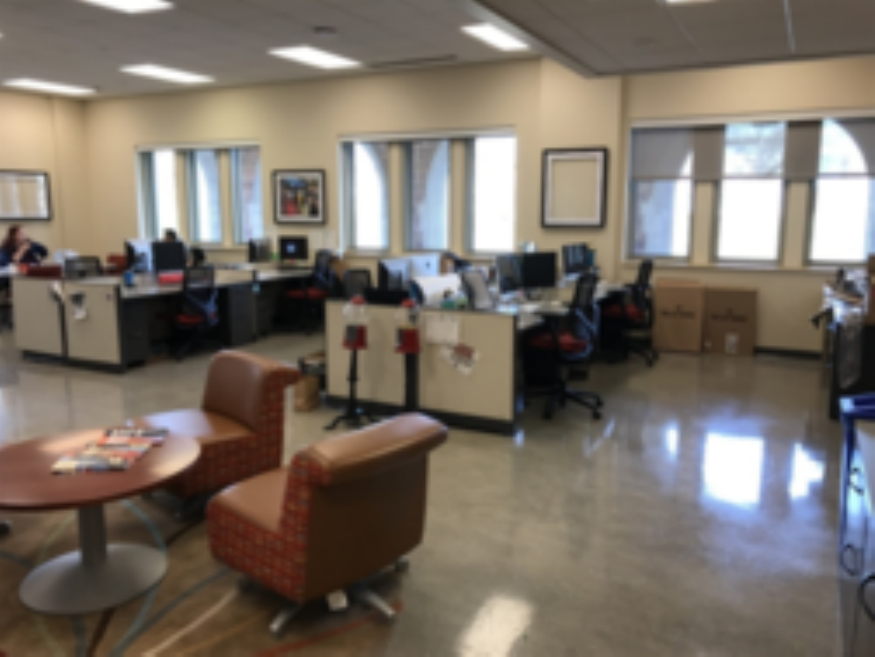} & 
    \includegraphics[width=0.242\linewidth]{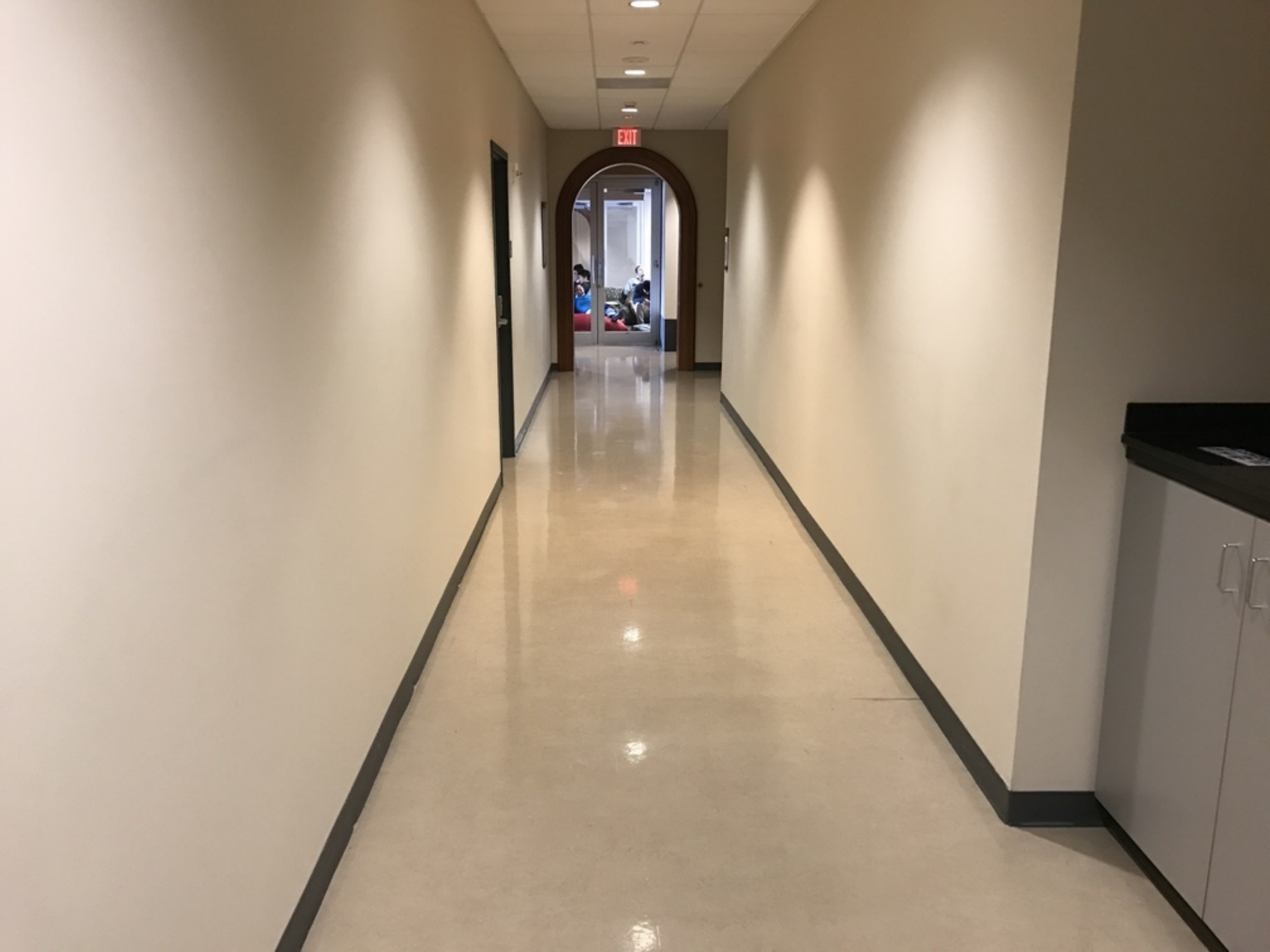} & 
    \includegraphics[width=0.242\linewidth]{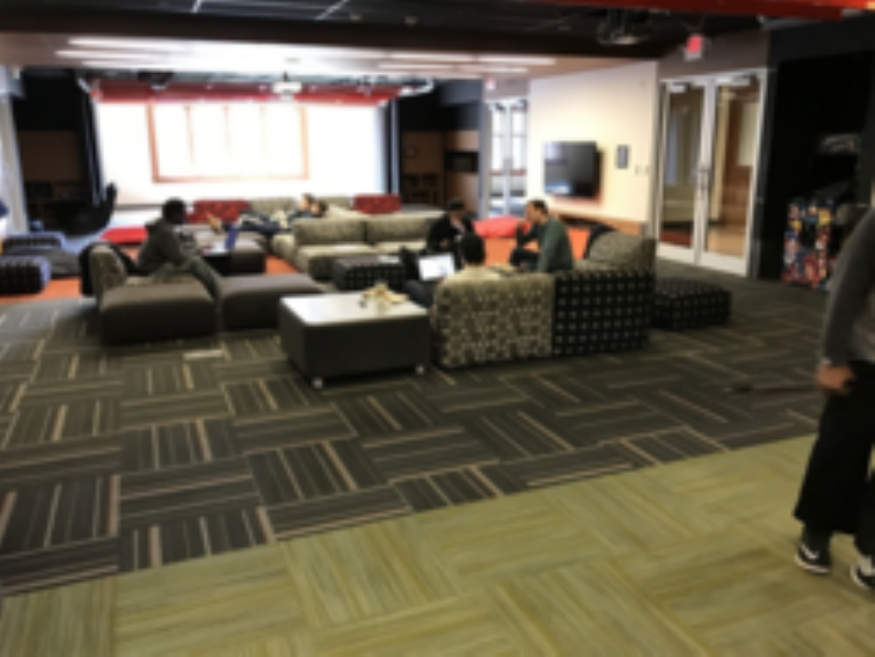} & 
    \includegraphics[width=0.242\linewidth]{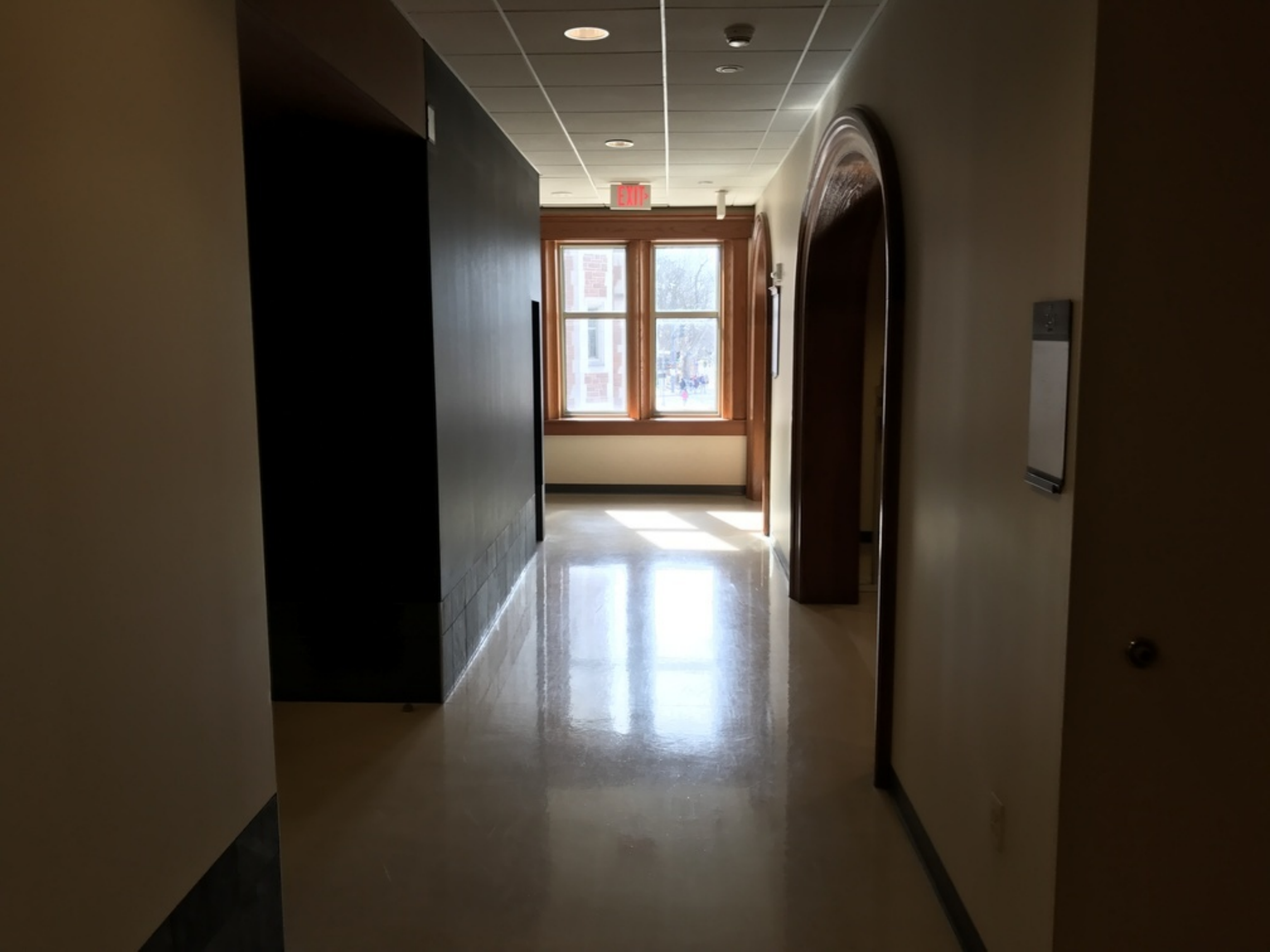} \\
    \end{tabular}
    \caption{{\bf Example images from InLoc dataset}. (Top) Database images. (Bottom) Query images. The selected images show the challenges encountered in indoor environments: even small changes in viewpoint lead to large differences in appearance; large textureless surfaces (\eg walls); self-repetitive structures (\eg corridors);  significant variation throughout the day due to different illumination sources (\eg, active vs.~indirect illumination). \label{fig:datasetimgsamples}}
    }
\end{figure}
}

\para{Query images.}
We captured 356 photos using a smart-phone camera (iPhone 7), distributed only across two floors, DUC1 and DUC2. The other three floors in the database are not represented in the query images, and play the role of confusers at search time, contributing to the building-scale localization scenario. Note that these query photos are taken at different times of the day, to capture the variety of occluders and layouts (\eg, people, furniture) as well as illumination changes.

{\tabcolsep = 1pt
\begin{figure}
    \centering
    {\footnotesize
    \begin{tabular}{cccc}
        \includegraphics[width=.242\linewidth]{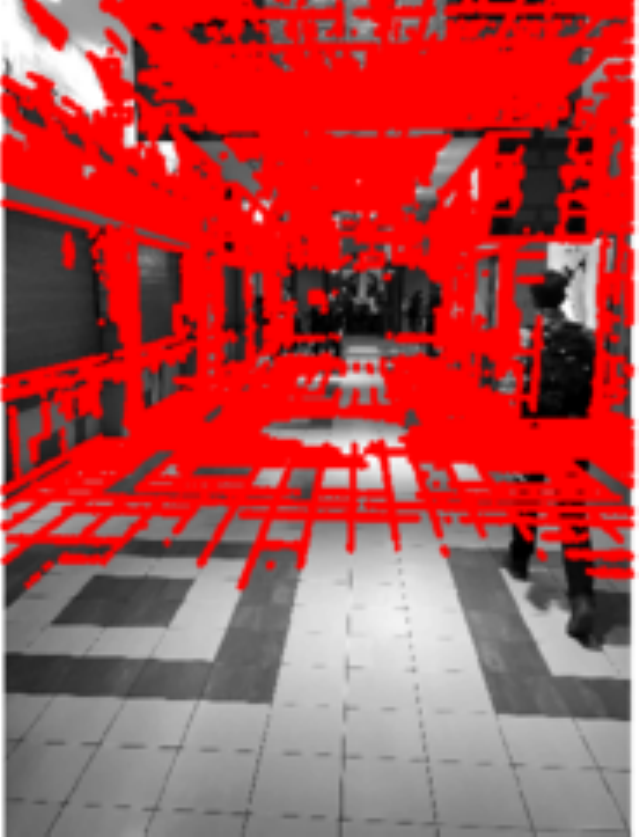} &
        \includegraphics[width=.242\linewidth]{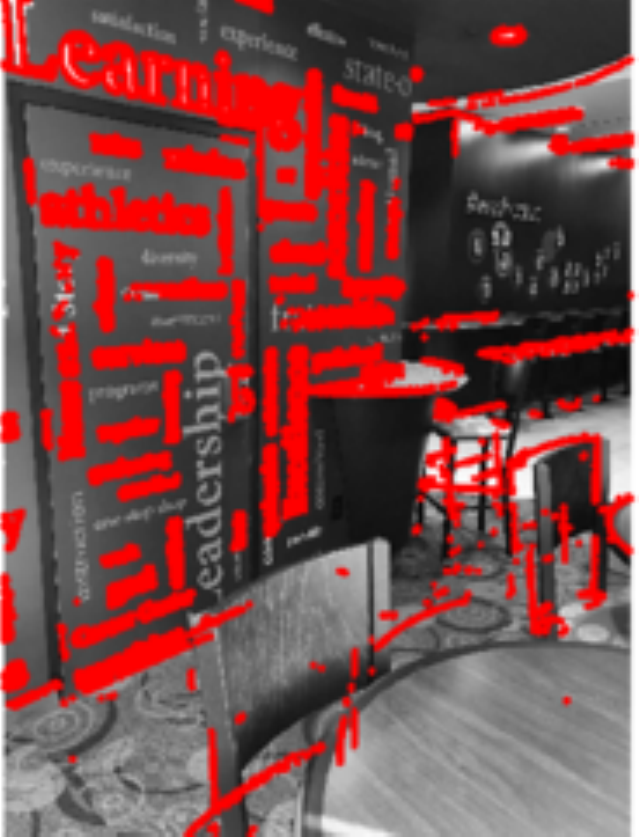} &
        \includegraphics[width=.242\linewidth]{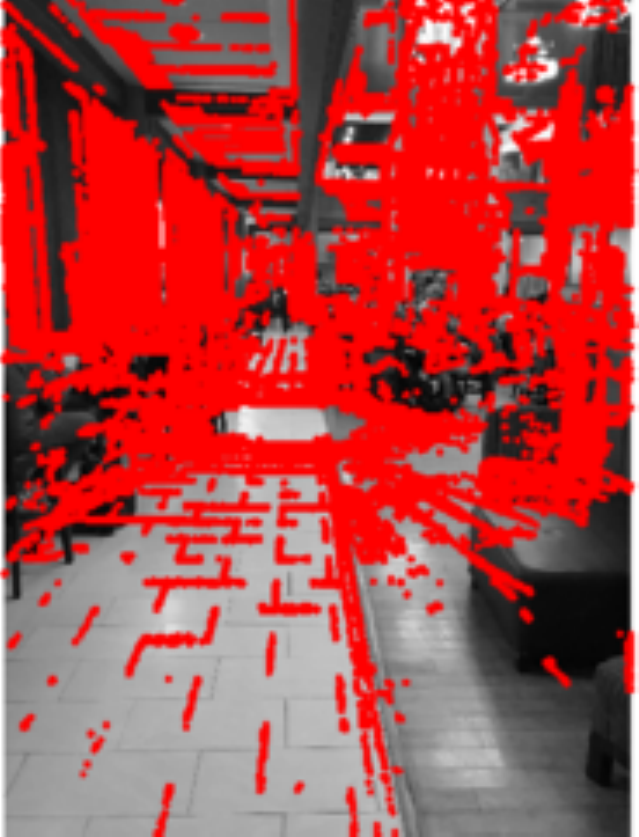} & 
        \includegraphics[width=.242\linewidth]{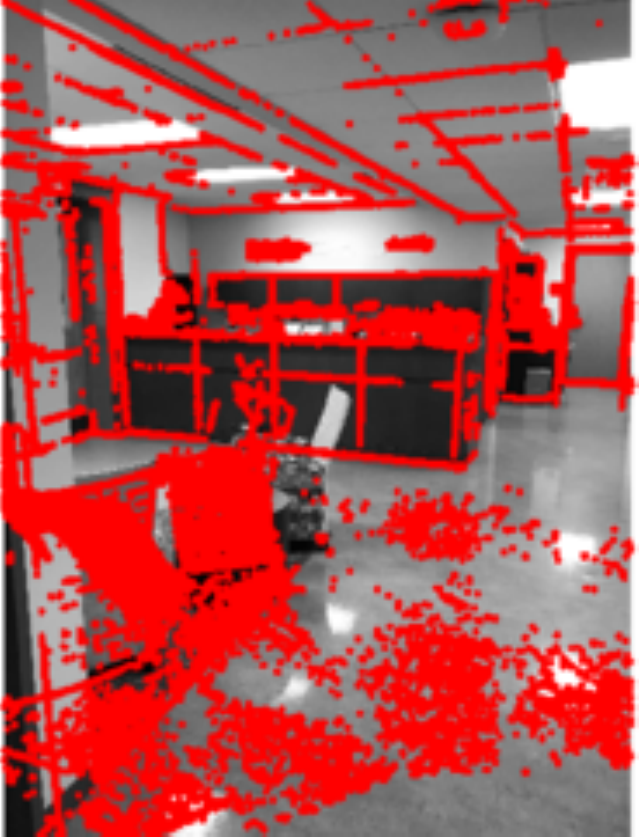}\\
    \end{tabular}
     \caption{{\bf Examples of verified query poses}. We evaluated the quality of the reference camera poses both visually and quantitatively, as described in section~\ref{sec:dataset}. 
     Red dots are the database 3D points projected onto a query image using its estimated pose. 
    }
    \label{fig:visualcheck}
    }
\end{figure}
}

\para{Reference pose generation.} 
For all query photos, we estimate 6DoF reference camera poses \wrt the 3D map. Each query camera reference pose is computed as follows:

\enum{(i) Selection of the visually most similar database images}. For each query, we manually select one panorama location which is visually most similar to the query image using the perspective images generated from the panorama.

\enum{(ii) Automatic matching of query images to selected database images}. We match the query and perspective images by using affine covariant features~\cite{mikolajczyk2004scale} and nearest-neighbor search followed by Lowe's ratio test~\cite{Lowe04}. 

\enum{(iii) Computing the query camera pose and visually verifying the reprojection}. All the panoramas (and perspective images) are already registered to the floor plan and have pixel-wise depth information. Therefore, we compute query pose via P3P-RANSAC~\cite{fischler1981random}, followed by bundle adjustment~\cite{ceres-solver}, using correspondences between query image points and scene 3D points obtained by feature matching. We evaluate the obtained poses visually 
by inspecting the reprojection of edges detected in the corresponding RGB panorama into the query image (see examples in figure~\ref{fig:visualcheck}). 
  
\enum{(iv) Manual matching of difficult queries to selected database images}. Pose estimation from automatic matches often gives inaccurate poses for difficult queries which are, \eg, far from any database image. Hence, for queries with significant misalignment in reprojected edges, we manually annotate 5 to 20 correspondences between image pixels and 3D points and apply step~(iii) on the manual matches.  
  
\enum{(v) Quantitative and visual inspection}. For all estimated poses, we measure the median reprojection error, computed as the distance of the reprojected 3D database point to the nearest edge pixel detected in the query image, after removing correspondences with gross errors (with distance over 20 pixels) due to, \eg, occlusions. For query images that have under 5 pixels median reprojection error, we manually inspect the reprojected edges in the query image and finally accept {\bf 329 reference poses} out of the 356 query images.

\section{Indoor visual localization with dense matching and view synthesis}
\noindent 
We propose a new method for large-scale indoor visual localization. 
We address the three main challenges of indoor environments: 

\para{(1) Lack of sparse local features}. Indoor environments are full of large textureless areas, \eg, walls, ceilings, floors and windows, where sparse feature extraction methods detect very few features. To overcome this problem, we use {\em multi-scale dense CNN features} for both image description and feature matching. Our features are generic enough to be pre-trained beforehand on (outdoor) scenes, avoiding costly re-training, \eg, as in~\cite{Brachmann2017DSAC,kendall2017geometric,Walch2017ICCV}, of the localization machine for each particular environment. 

\para{(2) Large image changes}. Indoor environments are cluttered with movable objects, \eg, furniture and people, and 3D structures, \eg, pillars add concave bays, causing severe occlusions when viewed from a close distance. The most similar images obtained by retrieval may therefore be visually very different from a query image. To overcome this problem, we rely on {\em dense feature matches to collect as much positive evidence as possible}. We employ image descriptors extracted from a convolutional neural network that can match higher-level structures of the scene rather than relying on matching individual local features. In detail, our pose estimation step performs coarse-to-fine dense feature matching, followed by geometric verification and estimation of the camera pose using P3P-RANSAC. 

\para{(3) Self-similarity}. Indoor environments are often very self-similar, \eg, due to many symmetric and repetitive elements on a large and small scale (corridors, rooms, tiles, windows, chairs, doors, \etc). Existing matching strategies count the positive evidence, \ie, how much of the image (or how many inliers) have been matched, to decide whether two images match.  This is, however, problematic as large textureless areas can be matched well, hence providing strong (incorrect) positive evidence.
To overcome this problem, we propose to count also the {\em negative evidence}, \ie, what portion of the image does not match, to decide whether two views are taken from the same location. To achieve this, we perform {\em explicit pose estimate verification based on view synthesis}. In detail, we compare the query image with a virtual view of the 3D model rendered from the estimated camera pose of the query. This novel approach takes advantage of the high quality of the RGBD image database and incorporates both the positive and negative evidence by counting matching and non-matching pixels across the entire query image. As shown by our experiments, this approach is orthogonal to the choice of local descriptors. The proposed verification by view synthesis is consistently showing a significant improvement regardless of the choice of features used for estimating the pose.

The pipeline of InLoc has the following three steps. Given a query image, (1) we obtain a set of candidate images by finding the N best  matching images from the reference image database registered to the map. (2) For these N retrieved candidate images, we compute the query poses using the associated 3D information that is stored together with the database images. (3) Finally, we re-rank the computed camera poses based on verification by view synthesis. The three steps are detailed next.

\newcommand{\thiswidth}{0.162\linewidth} 
{\tabcolsep=1pt
\begin{figure*}
    \centering
    {\footnotesize
    \begin{tabular}{cc|cc|cc}
    \includegraphics[width=\thiswidth]{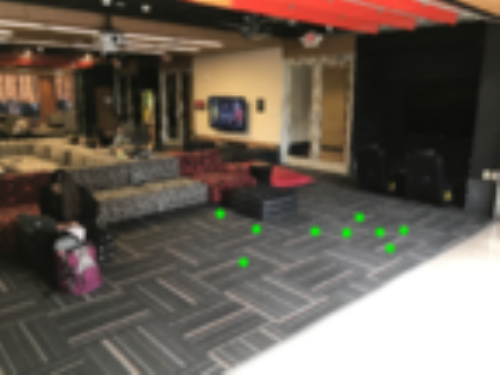} & 
    \includegraphics[width=\thiswidth]{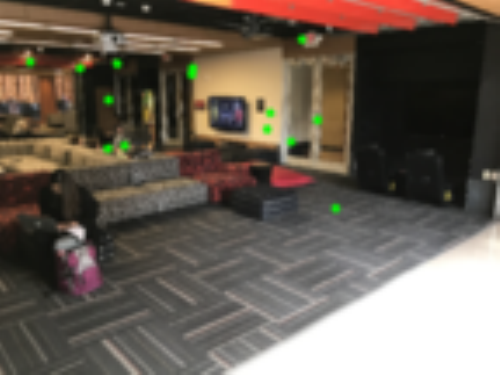} & 
    \includegraphics[width=\thiswidth]{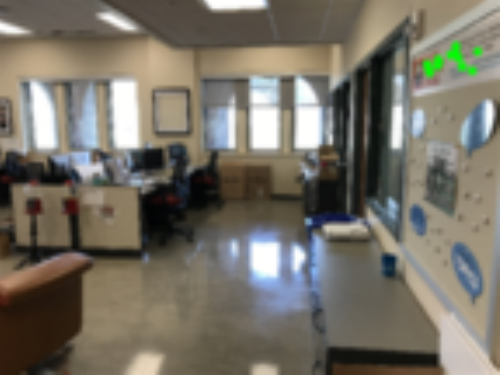} & 
    \includegraphics[width=\thiswidth]{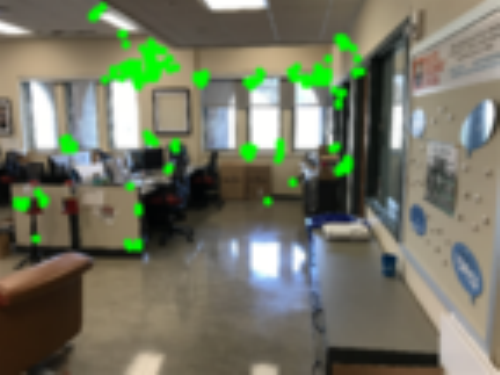} & 
    \includegraphics[width=\thiswidth]{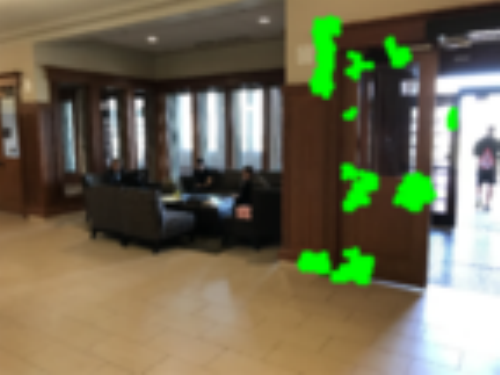} & 
    \includegraphics[width=\thiswidth]{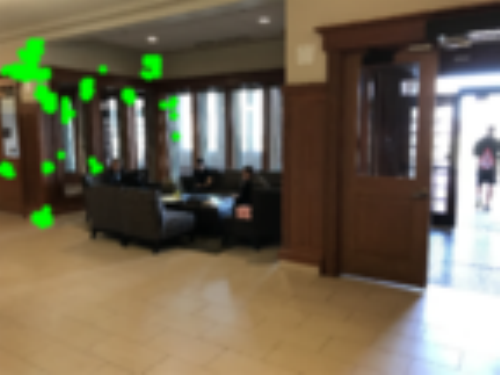} \\[-1pt]
    \includegraphics[width=\thiswidth]{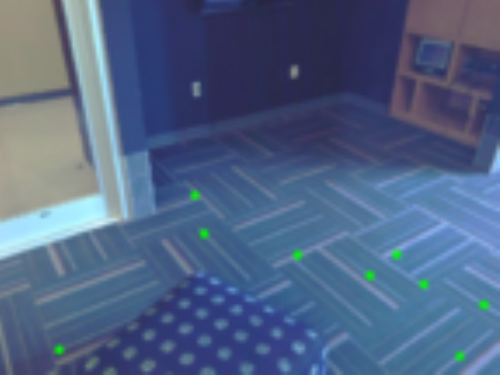} & 
    \includegraphics[width=\thiswidth]{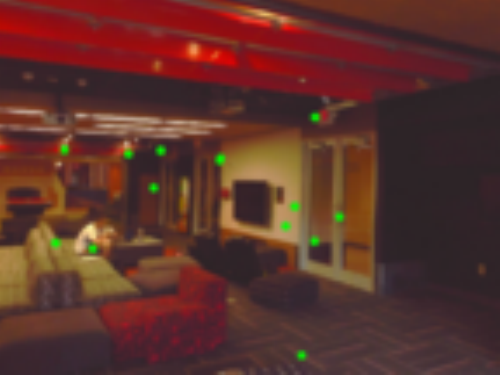} & 
    \includegraphics[width=\thiswidth]{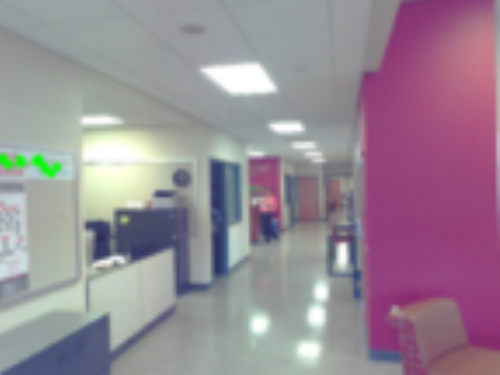} & 
    \includegraphics[width=\thiswidth]{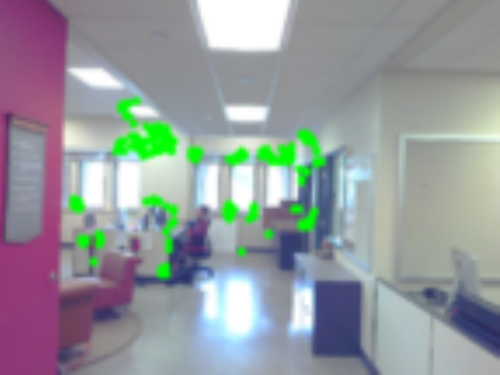} & 
    \includegraphics[width=\thiswidth]{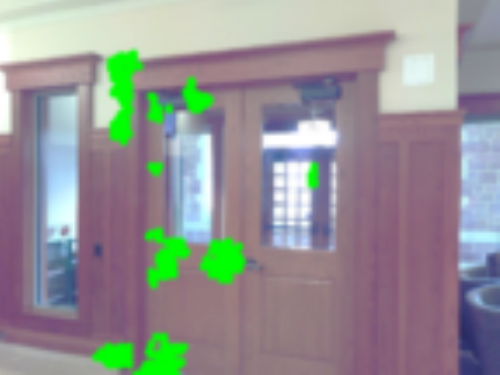} & 
    \includegraphics[width=\thiswidth]{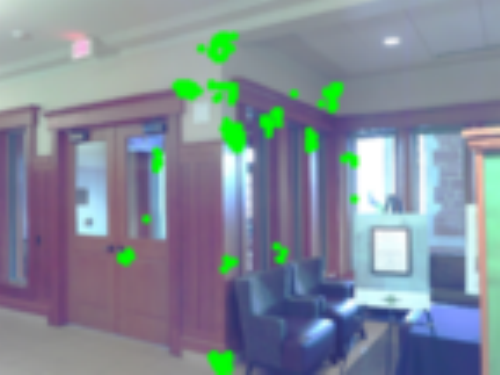} \\[-1pt]
    \includegraphics[width=\thiswidth]{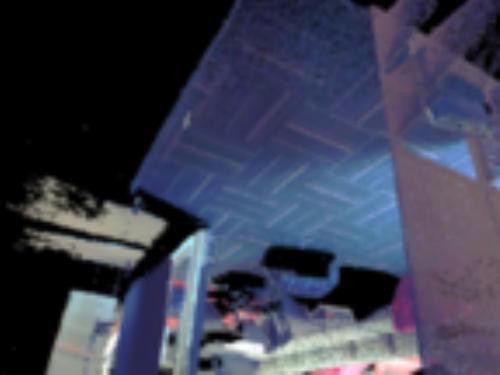} & 
    \includegraphics[width=\thiswidth]{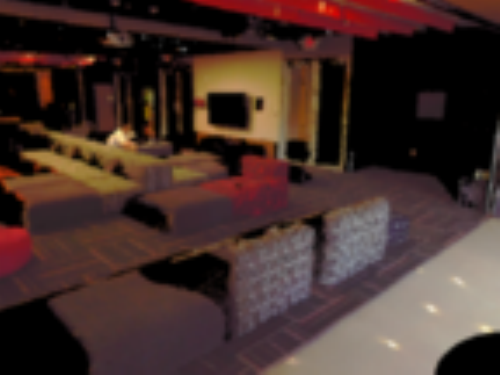} & 
    \includegraphics[width=\thiswidth]{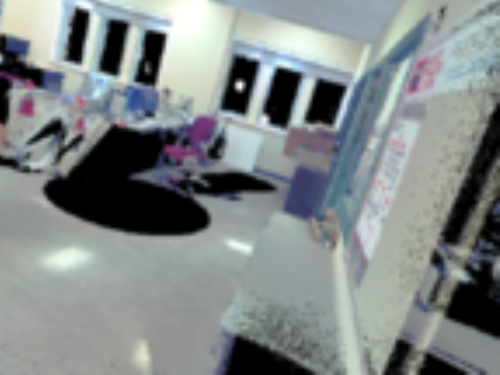} & 
    \includegraphics[width=\thiswidth]{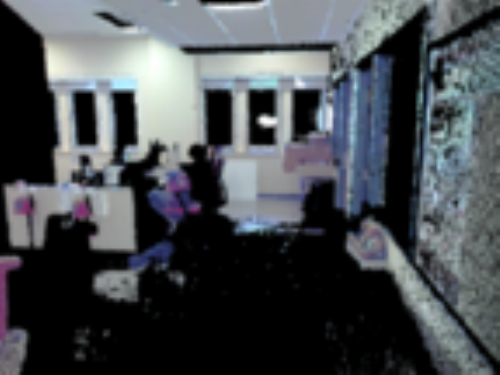} & 
    \includegraphics[width=\thiswidth]{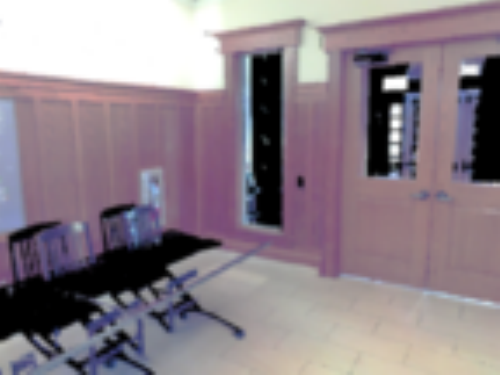} & 
    \includegraphics[width=\thiswidth]{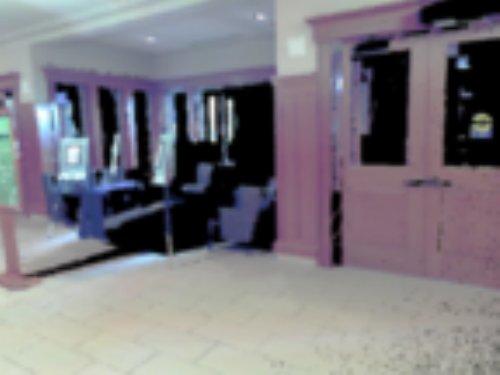} \\[-1pt]
    \includegraphics[width=\thiswidth]{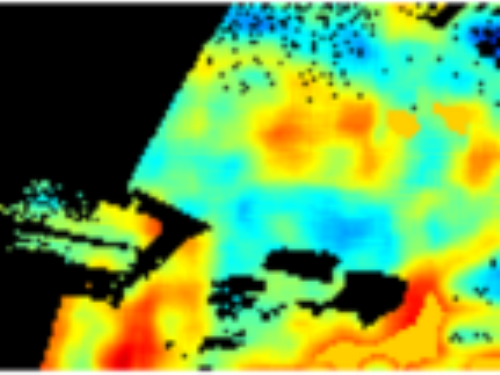} & 
    \includegraphics[width=\thiswidth]{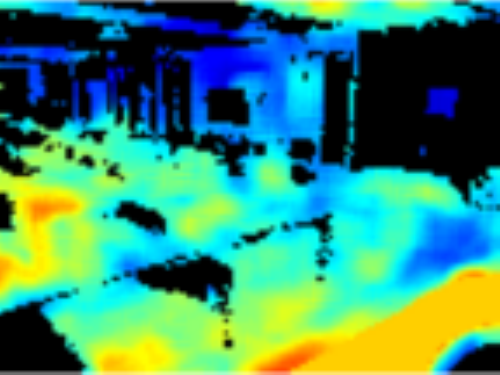} & 
    \includegraphics[width=\thiswidth]{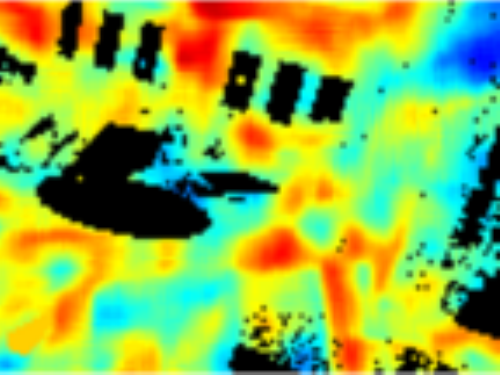} & 
    \includegraphics[width=\thiswidth]{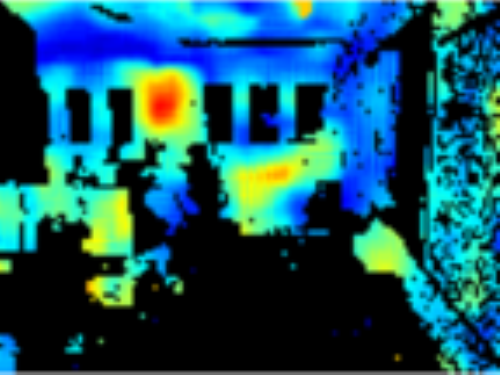} & 
    \includegraphics[width=\thiswidth]{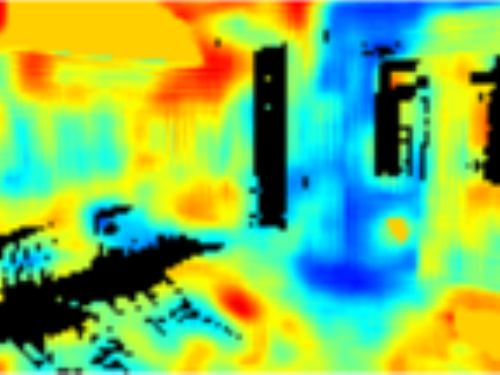} & 
    \includegraphics[width=\thiswidth]{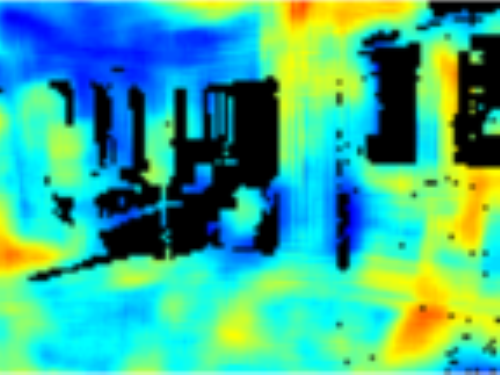} \\[-1pt] 
    8.39, 152.74$^{\circ}$ & 0.43, 2.05$^{\circ}$ & 0.27, 17.43$^{\circ}$ & 0.20, 0.72$^{\circ}$ & 7.97, 2.04$^{\circ}$ & 0.13, 1.95$^{\circ}$ \\[1pt] \hline
    Disloc~\cite{arandjelovic2014dislocation} & 
    NetVLAD~\cite{Arandjelovic16} & 
    NetVLAD~\cite{Arandjelovic16}+ & 
    NetVLAD~\cite{Arandjelovic16}+ & 
    NetVLAD~\cite{Arandjelovic16}+ & 
    {\bf InLoc:} NetVLAD~\cite{Arandjelovic16}+ \\
    ~ &
    ~ &
    SparsePE & 
    {\bf DensePE} & 
    {\bf DensePE} & 
    {\bf DensePE}+{\bf DensePV}\\
    \end{tabular}
    \caption{{\bf Qualitative comparison of different localization methods (columns).} From top to bottom: query image, the best matching database image, synthesized view at the estimated pose (without inter/extra-polation), error map between the query image and the synthesized view, localization error (meters, degrees). Green dots are the inlier matches obtained by P3P-LO-RANSAC. Methods using the proposed dense pose estimation (DensePE) and dense pose verification (DensePV) are shown in bold. The query images in the 2nd, 4th and 6th column are well localized within 1.0 meters and 5.0 degrees whereas localization results in the 1st, 3rd and 5th column are incorrect. 
    \label{fig:qualitative}}
    }
\end{figure*}
}

\subsection{Candidate pose retrieval \label{sec:candidate_retrieval}}
\noindent 
As demonstrated by existing work~\cite{torii201524,Arandjelovic16,kim2017learned}, 
aggregating feature descriptors computed densely on a regular grid mitigates issues such as a lack of repeatability of local features detected on textureless scenes, large-illumination changes, and a lack of discriminability of image description,
dominated by features from repetitive structures (burstiness). 
As already mentioned in section~\ref{sec:intro}, these problems are also occurring in large-scale indoor localization, which motivates our choice of using an image descriptor based on dense feature aggregation. Both query and database images are described by NetVLAD~\cite{Arandjelovic16} (but other variants could also be used), normalized L2 distances of the descriptors are computed, and the poses of the N best matching images from the database are chosen as candidate poses. In section~\ref{sec:exp}, we compare our approach with the state-of-the-art image descriptors based on local feature detection and show benefits of our approach for indoor localization. 

\subsection{Pose estimation using dense matching \label{sec:pose_estimation}}
\noindent 
A severe problem in indoor localization is that standard geometric verification based on local feature detection~\cite{philbin2007object,sattler2016large} does not work on textureless or self-repetitive scenes, such as corridors, where robots (and also humans) often get lost. Motivated by the improvements in candidate pose retrieval with dense feature aggregation (Section~\ref{sec:candidate_retrieval}), we use features  densely extracted on a regular grid for verifying and re-ranking the candidate images by feature matching and pose estimation. A possible approach would be to match DenseSIFT~\cite{liu2008sift} followed by RANSAC-based verification. Instead of tailoring DenseSIFT description parameters (patch sizes, strides, scales) to match across images with significant viewpoint changes, we use an image representation extracted by a convolutional neural network (VGG-16~\cite{simonyan2014very}) as a set of multi-scale features extracted on a regular grid that describes more higher-level information with a larger receptive field (patch size). 

We first find geometrically consistent sets of correspondences using the coarser 
conv5 layer containing high-level information.
Then we refine the correspondence by searching for additional matches on the conv3 layer. 
Examples in figure~\ref{fig:qualitative} demonstrate that our dense CNN matching (4th column) obtains better matches in indoor environments when compared to matching standard local features (3rd column), even for less-textured areas. 
Notice that dense-feature extraction and description requires no additional computation at query time as the intermediate convolutional layers are already computed when extracting the NetVLAD descriptors as described in section~\ref{sec:candidate_retrieval}. As will also be demonstrated in section~\ref{sec:exp}, memory requirements and computational speed of feature matching can be addressed by binarizing the convolutional features without loss in matching performance.

As perspective images in our database have depth values, and hence associated 3D points, the query camera pose can be estimated by finding pixel-to-pixel correspondences between the query and the matching database image followed by P3P-RANSAC~\cite{fischler1981random}. 

\subsection{Pose verification with view synthesis \label{sec:pose_verification}}
\noindent 
We propose here to collect both positive and negative evidence to determine what {\em is} and {\em is not} matched\footnote{The impact of negative evidence in feature aggregation is demonstrated in~\cite{Jegou12}.}. This is achieved by harnessing the power of the high-quality RGBD image database that provides a dense and accurate 3D structure of the indoor environment. 
This structure is used to render a virtual view that shows how the scene would look like from the estimated query pose. The rendered image enables us to count, in a pixel-wise manner, both positive and negative evidence by counting which regions are and are not consistent between the query image and the underlying 3D structure. To gain invariance to illumination changes and small misalignments, we evaluate image similarity by comparing local patch descriptors (DenseRootSIFT~\cite{liu2008sift,arandjelovic2012three}) at corresponding pixel locations. 
The final similarity is computed as the median of descriptor distances across the entire image while ignoring areas with missing 3D structure. 

\section{Experiments \label{sec:exp}}
\noindent
We first describe the experimental setup for evaluating visual localization performance using our dataset (Section~\ref{subsec:exsetup}). The proposed method, termed ``InLoc'', is compared with  state-of-the-art methods (Section~\ref{subsec:baselines}) and we show the benefits of each component in detail (Section~\ref{subsec:eachstep}). 

\subsection{Implementation details \label{subsec:exsetup}}
\noindent 
In the candidate pose retrieval step, we retrieve 100 candidate database images using NetVLAD. We use the implementation provided by the authors and the pre-trained Pitts30K~\cite{Arandjelovic16} VGG-16~\cite{simonyan2014very} model to generate $4,096$-dimensional NetVLAD descriptor vectors.  

In the second pose estimation step, we obtain tentative correspondences by matching densely extracted convolutional features in a coarse-to-fine manner: we first find mutually nearest matches among the conv5 features and then find matches in the finer conv3 features restricted by the coarse conv5 correspondences. The tentative matches are geometrically verified by estimating up to two homographies using RANSAC~\cite{fischler1981random}. 
We re-rank the 100 candidates using the number of RANSAC inliers and keep the top-10 database images. For each of the 10 images, the 6DoF query pose is computed by P3P-LO-RANSAC~\cite{lebeda2012fixing} (referred to as \emph{DensePE}), assuming a known focal length, \eg, from EXIF data, using the inlier matches and depth (\ie the 3D structure)  associated to each database image.

In the final pose verification step, we generate synthesized views by rendering colored 3D points while taking care of self-occlusions. For computing the scores that measure the similarities of the query image and the image rendered from the estimated pose, we use the DenseSIFT extractor and its RootSIFT descriptor~\cite{liu2008sift,arandjelovic2012three} from VLFeat~\cite{Vedaldi10a}\footnote{When computing the descriptors, the blank pixels induced by missing 3D points are filled by linear inter(/extra)-polation using the values of non-blank pixels on the boundary. }. Finally, we localize the query image by the best pose among its top-10 candidates. 

\para{Evaluation metrics. }
We evaluate the localization accuracy as the  consistency of the estimated poses with our reference poses. We measure positional and angular differences in meters and degrees between the estimated poses and the manually verified reference poses.

\subsection{Comparison with the state-of-the-art methods \label{subsec:baselines}}

\para{Direct 2D-3D matching~\cite{sattler2017efficient,sattler2015hyperpoints}.} We first compare with a variation\footnote{Due to the sparse sampling of viewpoints in our indoor dataset, we cannot establish feature tracks between database images. This prevents us from applying algorithms relying on co-visibility~\cite{sattler2017efficient,li2012worldwide,choudhary2012visibility,Zeisl2015ICCV,sattler2015hyperpoints}.} of a state-of-the-art 3D structure-based image localization approach~\cite{sattler2015hyperpoints}. 
We compute affine covariant RootSIFT features for all the database images and associate them with 3D coordinates via the known scene geometry. Features extracted from a query image are then matched to the database 3D descriptors~\cite{Muja09}. 
We select at most five database images receiving the largest numbers of matches and use all these matches together for pose estimation. Similar to~\cite{sattler2015hyperpoints}, we did not apply Lowe's ratio test~\cite{Lowe04} as it lowered the performance. The 6DoF query pose is finally computed by P3P-LO-RANSAC~\cite{lebeda2012fixing}. 
As shown in table~\ref{tab:baselines}, InLoc outperforms direct 2D-3D matching by a large margin ($40.7\%$ at the localization accuracy of 0.5m). We believe that this is because our large-scale indoor dataset involves many distractors and large viewpoint changes that present a major challenge for 3D structure-based methods. 

\begin{table}[t]
    \centering
    \setlength{\tabcolsep}{3pt}
    {\footnotesize 
    \begin{tabular}{r|cccc}
    ~ &
    Direct2D-3D & 
    Disloc~\cite{arandjelovic2014dislocation} &
    NetVLAD & 
    InLoc \\ 
    ~ &
    \cite{sattler2015hyperpoints} & 
    +SparsePE &
    +SparsePE & 
    (Ours)
    \\ \hline
     0.25m & 11.9 & 20.1 & 21.3 & {\bf 38.9}  \\
     0.50m & 15.8 & 29.5 & 30.7 & {\bf 56.5} \\  
     1.00m & 22.5 & 41.0 & 42.6 & {\bf 69.9} \\  
    \end{tabular}
    }
    {\scriptsize 
    \caption{{\bf Comparison with the state-of-the-art localization methods on the InLoc dataset.} We show the rate (\%) of correctly localized queries within a given distance (m) threshold and within a $10^{\circ}$ angular error threshold.}
    \label{tab:baselines}
    }
\end{table}

{\tabcolsep=1pt
\begin{figure*}[t]
    \centering
    {\footnotesize
    \begin{tabular}{ccc}
    \includegraphics[width=0.4\linewidth]{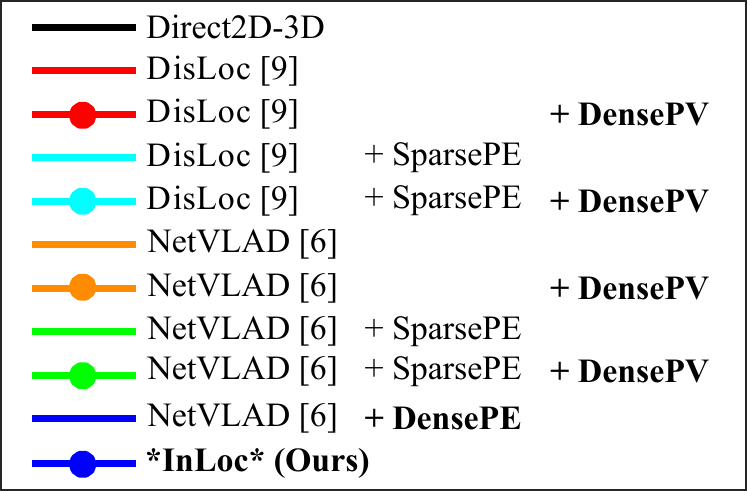} & 
    \includegraphics[width=0.292\linewidth]{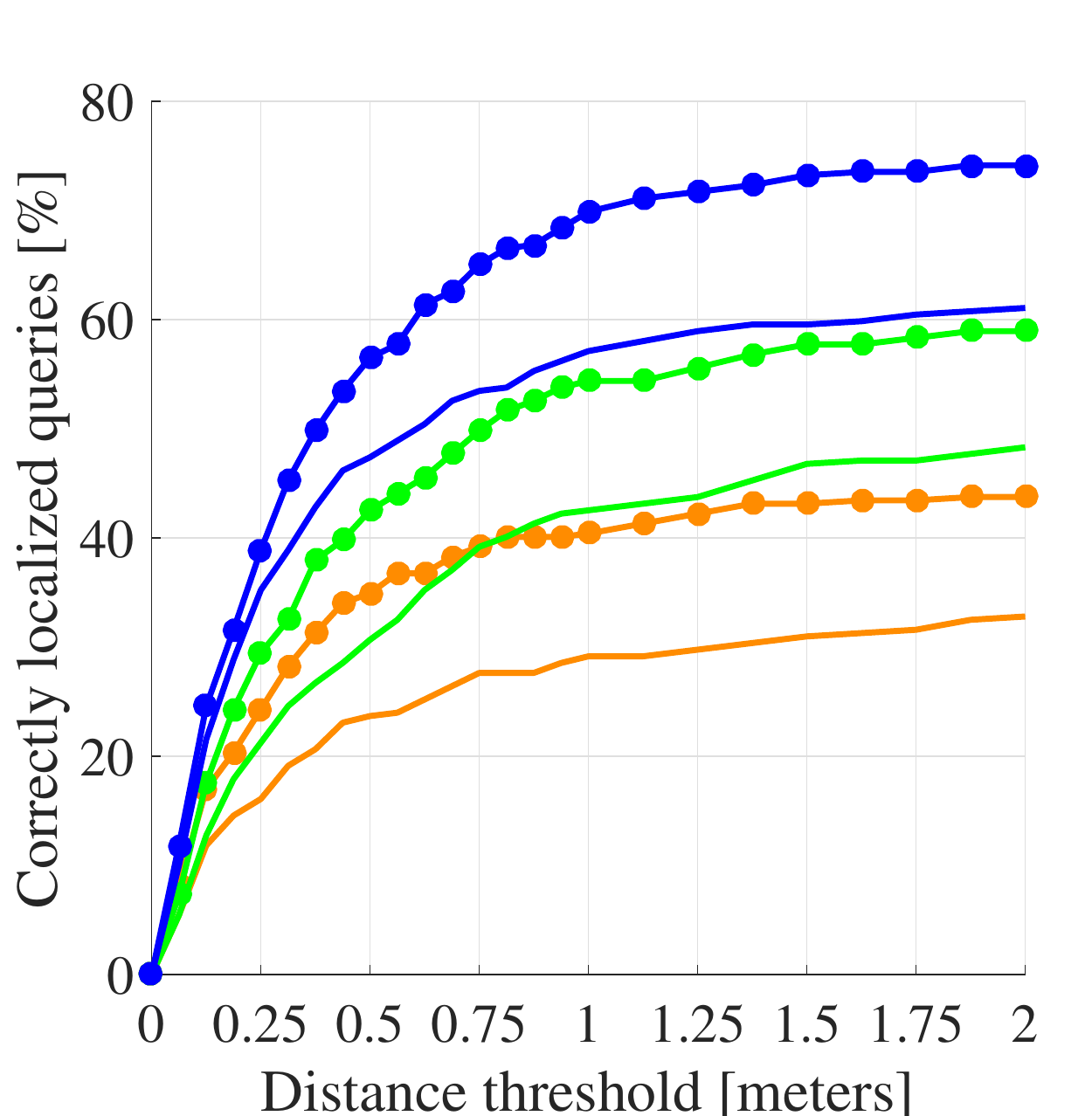} &  \includegraphics[width=0.292\linewidth]{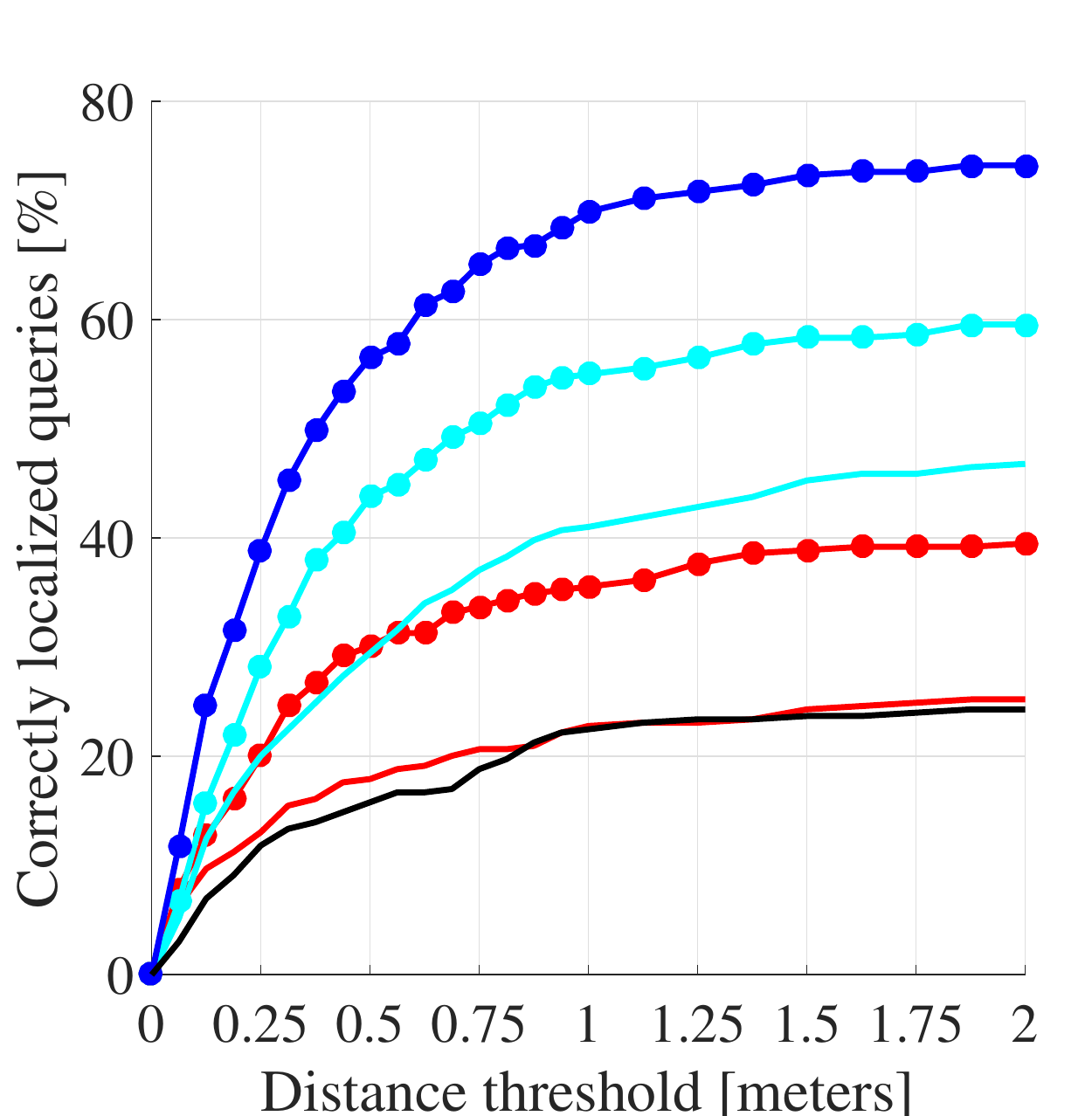}\\[2pt]
    ~ & (a) NetVLAD baselines & (b) Other baselines\\[2pt]
    \end{tabular}
    \caption{{\bf Impact of different components.} The graphs show impact of dense matching ({\bf DensePE}) and dense pose verification ({\bf DensePV}) on pose estimation quality for (a) the pose candidates retrieved by NetVLAD and (b) state-of-the-art baselines. Plots show the fraction of correctly localized queries (y-axis) within a certain distance (x-axis) whose rotation error is at most  $10^{\circ}$.  \label{fig:exrecall2d}
    }
    }
\end{figure*}
}

\para{Disloc~\cite{arandjelovic2014dislocation} + sparse pose estimation (SparsePE)~\cite{philbin2007object}.} We next compare with the state-of-the-art image retrieval-based localization method. Disloc represents images using bag-of-visual-words with Hamming-Embedding~\cite{jegou2008hamming} 
while also taking local descriptor space density into account. 
We use a publicly available implementation~\cite{sattler2016large} of Disloc with a 200K vocabulary trained on affine covariant features~\cite{mikolajczyk2004scale}, described by RootSIFT~\cite{arandjelovic2012three}, extracted from the database images of our indoor dataset. 
The top-100 candidate images shortlisted by Disloc are re-ranked by spatial verification~\cite{philbin2007object} using (sparse) affine covariant features~\cite{mikolajczyk2004scale}. 
The ratio test~\cite{Lowe04} was not applied here as it was removing too many features that need to be retained in the indoor scenario. 
Using the inliers, the 6DoF query pose is computed with P3P-LO-RANSAC~\cite{lebeda2012fixing}. To make a fair comparison, we use exactly the same features and P3P-LO-RANSAC for pose estimation as the direct 2D-3D matching method described above.  
As shown in table~\ref{tab:baselines}, Disloc~\cite{arandjelovic2014dislocation}+SparsePE~\cite{philbin2007object} results in a $13.7\%$ performance gain compared to Direct 2D-3D matching~\cite{sattler2017efficient}. This can be attributed to the image retrieval step that discounts burst of repetitive features. However, the results are still significantly worse compared to our InLoc approach. 

\para{NetVLAD~\cite{Arandjelovic16} + sparse pose estimation (SparsePE)~\cite{philbin2007object}.} 
We also evaluate a variation of the above image retrieval-based localization method. Here the candidate shortlist is obtained by NetVLAD~\cite{Arandjelovic16}, which is then re-ranked using SparsePE~\cite{philbin2007object}, followed by pose estimation using P3P-LO-RANSAC~\cite{lebeda2012fixing}. This is a strong baseline building on the state-of-the-art place recognition results obtained by~\cite{Arandjelovic16}.
Interestingly, as shown in table~\ref{tab:baselines}, there is no significant difference between NetVLAD+SparsePE and DisLoc+SparsePE, which is in line with results reported in outdoor settings~\cite{sattler2017large}. 
Yet, NetVLAD outperforms DisLoc ($5.8\%$ at the localization accuracy of 0.5m) before re-ranking via SparsePE (\cf figure~\ref{fig:exrecall2d}) in this indoor setting (see also figure~\ref{fig:qualitative}). 
Overall, both methods, even though they represent the state-of-the-art in outdoor localization, still perform significantly worse than our proposed approach based on  dense feature matching and view synthesis. 

\subsection{Evaluation of each component \label{subsec:eachstep}} 
\noindent 
Next, we demonstrate the benefits of the  individual components of our approach.

\para{Benefits of pose estimation using dense matching.}
Using the NetVLAD retrieval as the base retrieval method (Figure~\ref{fig:exrecall2d}~(a)),
our pose estimation with dense matching (NetVLAD~\cite{Arandjelovic16}+{\bf DensePE}~(blue line)) constantly improves the localization rate by about $15\%$ when compared to the state-of-the-art sparse local feature matching (NetVLAD~\cite{Arandjelovic16}+SparsePE~(green line)). This result supports our conclusion that dense feature matching and verification is superior to sparse feature matching for often weakly textured indoor scenes. 
This effect is also clearly demonstrated in qualitative results in figure~\ref{fig:qualitative} (cf. columns 3 and 4).

\para{Benefits of pose verification with view synthesis.}
We apply our pose verification step ({\bf DensePV}) to the top--10 pose estimates obtained by different spatial re-reranking methods. Results are shown in figure~\ref{fig:exrecall2d} and demonstrate significant and consistent improvements obtained by our pose verification approach (compare ``-$\bullet$-'' to ``---'' in figure~\ref{fig:exrecall2d}). Improvements are most pronounced for the position accuracy within 1.5 meters (13\% or more). 

\para{Binarized representation. }
A binary representation (instead of floats) of features in the intermediate CNN layers significantly reduces memory requirements. 
We use feature binarization that follows the standard Hamming embedding approach~\cite{jegou2008hamming} but without dimensionality reduction. Matching is then performed by computing Hamming distances. This simple binarization scheme results in a negligible performance loss (less than 1\% at 0.5 meters) compared to the original descriptors, which is in line with results reported for object recognition~\cite{AgrawalGM14}. 
At the same time, binarization reduces the memory requirements by a factor of 32, compressing 428GB of original descriptors to just 13.4GB. 

\para{Comparison with learning based localization methods.} 
We have attempted a comparison with DSAC~\cite{Brachmann2017DSAC}, which is a  state-of-the-art pose estimator for indoor scenes. Despite our best efforts, training DSAC on our indoor dataset failed to converge. We believe this is because the RGBD scans in our database are sparsely distributed~\cite{wijmans17rgbd} and each scan has only a small overlap with neighboring scans. Training on such a dataset is challenging for methods designed for densely captured RGBD sequences~\cite{glocker2013real}. We believe this would also be the case for PoseNet~\cite{kendall2017geometric}, another method for CNN-based pose regression. We do provide the comparison with DSAC and PoseNet on much smaller datasets next. 

\begin{table}[tb]
    \centering
    \setlength{\tabcolsep}{2.7pt}
    {\footnotesize
    \begin{tabular}{r|ccccc}
    ~ &
    Disloc~\cite{arandjelovic2014dislocation} & 
    NetVLAD~\cite{Arandjelovic16} & 
    NetVLAD~\cite{Arandjelovic16} & 
    InLoc \\ 
    ~&
    +SparsePE & 
    +SparsePE &
    +DensePE & 
    (Ours)
    \\ \hline
    90 bldgs. & 
    0.42, 4.58$^{\circ}$ & 0.44, 4.70$^{\circ}$ & 0.23, 2.53$^{\circ}$ & {\bf 0.17, 2.15$^{\circ}$} \\    
    \end{tabular}
    }
    {\scriptsize 
    \caption{{\bf Comparison on Matterport3D~\cite{Chang20173DV}.} Numbers show the median positional (m) and angular (degrees) errors. 
    }
    \label{tab:matterport}
    }
        \vspace*{-2mm}
\end{table}
\begin{table}[t]
    \centering
    \setlength{\tabcolsep}{2.7pt}
    {\scriptsize
    \begin{tabular}{c|ccccc}
    ~ & 
    PoseNet & 
    ActiveSearch  & 
    DSAC & 
    NetVLAD~\cite{Arandjelovic16} & 
    NetVLAD~\cite{Arandjelovic16} \\ 
    Scene & 
    \cite{kendall2017geometric} & 
    \cite{sattler2017efficient}  & 
    \cite{Brachmann2017DSAC,brachmann2017learning} & 
    +SparsePE~\cite{philbin2007object} & 
    +{\bf DensePE} \\ \hline
    Chess & 
    13, 4.48$^{\circ}$ & 4, 1.96$^{\circ}$ & {\bf 2}, 1.2$^{\circ}$ & 4, 1.83 & 3, {\bf 1.05$^{\circ}$} \\
    Fire & 
    27, 11.3$^{\circ}$ & \textbf{3}, 1.53$^{\circ}$ & 4, 1.5$^{\circ}$ & 4, 1.55 & {\bf 3}, {\bf 1.07$^{\circ}$} \\
    Heads 
    & 17, 13.0$^{\circ}$ & {\bf 2}, 1.45$^{\circ}$ & 3, 2.7$^{\circ}$ & {\bf 2}, 1.65 & {\bf 2}, {\bf 1.16$^{\circ}$} \\
    Office 
    & 19, 5.55$^{\circ}$ & 9, 3.61$^{\circ}$ & 4, 1.6$^{\circ}$ & 5, 1.49 & {\bf 3}, {\bf 1.05$^{\circ}$} \\
    Pumpkin 
    & 26, 4.75$^{\circ}$ & 8, 3.10$^{\circ}$ & {\bf 5}, 2.0$^{\circ}$ & 7, 1.87 & {\bf 5}, {\bf 1.55$^{\circ}$} \\
    Red kit. 
    & 23, 5.35$^{\circ}$ & 7, 3.37$^{\circ}$ & 5, 2.0$^{\circ}$ & 5, 1.61 & {\bf 4}, {\bf 1.31$^{\circ}$} \\
    Stairs 
    & 35, 12.4$^{\circ}$ & {\bf 3}, {\bf 2.22}$^{\circ}$ & 117, 33.1$^{\circ}$     
    & 12, 3.41 & 9, 2.47$^{\circ}$ \\[2pt] 
    \end{tabular}
    }
    {\footnotesize 
    \caption{{\bf Evaluation on the 7 Scenes dataset~\cite{glocker2013real,shotton2013scene}}. Numbers show the median positional (cm) and angular errors (degrees). } 
    \label{tab:ex7scenes}
    }
\end{table}

\subsection{Evaluation on other datasets \label{subsec:7scenes}}
\noindent We also evaluate InLoc on two existing indoor datasets~\cite{Chang20173DV,shotton2013scene} to confirm the relevance of our results.
The Matterport3D~\cite{Chang20173DV} dataset consists of RGBD scans of 90 buildings. Each RGBD scan contains 18 images that capture the scene around the scan position with known camera poses. We created a test set by randomly choosing 10\% of the scan positions and selected their horizontal views. This resulted in 58,074 database images and a query set of 6,726 images. Results are shown in table~\ref{tab:matterport}. 
Our approach (InLoc) outperforms the baselines, which is in line with results on the InLoc dataset. 
We also tested PoseNet~\cite{kendall2017geometric} and DSAC~\cite{Brachmann2017DSAC} on a single (the largest) building. The test set is created in the same manner as above and contains 1,884 database images and 210 query images. 
Even in this much easier case, DSAC fails to converge. PoseNet produces large localization errors (24.8 meters and 80.0 degrees) in comparison with InLoc (0.26 meters and 2.78 degrees). 

We also report results on the 7 Scenes dataset~\cite{glocker2013real,shotton2013scene} which is, while relatively small, a standard benchmark for indoor localization. 
The 7 Scenes dataset~\cite{shotton2013scene} consists of geometrically-registered video frames representing seven scenes, together with associated depth images and camera poses. Table~\ref{tab:ex7scenes} shows localization results for our approach (NetVLAD+{\bf DensePE}) compared with state-of-the-art methods~\cite{sattler2017efficient,Brachmann2017DSAC,kendall2017geometric}. 
Note that our approach performs comparably to these methods on this relatively small and densely captured data, while it does not need any scene specific training (which is needed by~\cite{kendall2017geometric, Brachmann2017DSAC}).

\section{Conclusion}
\noindent
We have presented InLoc -- a new approach for large-scale indoor visual localization that estimates the 6DoF camera pose of a query image with respect to a large indoor 3D map. To overcome the difficulties of indoor camera pose estimation, we have developed new pose estimation and verification methods that use dense feature extraction and matching in a sequence of progressively stricter verification steps. 
The localization performance is evaluated on a new large indoor dataset with realistic and challenging query images captured by mobile phones. Our results demonstrate significant improvements  compared to  state-of-the-art localization methods. To encourage further progress on  high-accuracy large-scale indoor localization, we make our dataset publicly available~\cite{projectpage}.

\para{Acknowledgements.} This work was partially supported by 
 JSPS KAKENHI Grant Numbers 15H05313, 17H00744, 17J05908, EU-H2020 project LADIO~No.~731970, ERC grant LEAP No.\ 336845, CIFAR Learning in Machines $\&$ Brains program and the European Regional Development Fund under the project IMPACT (reg. no. CZ$.02.1.01/0.0/0.0/15\_003/0000468$). The authors would like to express the deepest appreciation to Yasutaka Furukawa for his arrangement to capture query photographs at Washington University in St. Louis.

\section*{Appendix}
\noindent This appendix first provides additional examples of query images and their reference poses in our \textbf{InLoc} dataset (section~\ref{sec:queryimages}). We also present additional qualitative results, illustrating situations in which the proposed \textbf{InLoc} method succeeds while the investigated baseline methods fail (section~\ref{sec:qual}).

\appendix
\renewcommand{\thetable}{\Alph{table}}
\renewcommand{\thefigure}{\Alph{figure}}
\setcounter{figure}{0}
\setcounter{table}{0}

\section{Additional examples of query images and reference poses in the InLoc dataset \label{sec:queryimages}}
\noindent
Figure~\ref{fig:refposes_all} shows the 3D maps (grey dots), the 329 reference poses of the query images (blue dots), and the 129 database scan positions (red circles) in our InLoc dataset. The query images are distributed across two floors (DUC1 and DUC2) that cover an area of $\approx$ 100,000 $ft^2$ (9,290\ $m^2$) each~\cite{wijmans17rgbd}, and are taken from significantly distant positions from database scans. 

Figure~\ref{fig:refposes_qual} illustrates the verification process for the reference poses. We show example query images on the 1st and 3rd row. The edges extracted on the best matching database image were reprojected on the query image (2nd and 4th row) to verify the quality of the reference poses. The manually and visually verified reference poses, in total 329, have at most 5 pixels median re-projection error, out of which, 101 reference poses have median re-projection error below 1 pixel. 

{\tabcolsep=1pt
\begin{figure*}
    \centering
    \begin{tabular}{c}
    \includegraphics[width=0.9\linewidth]{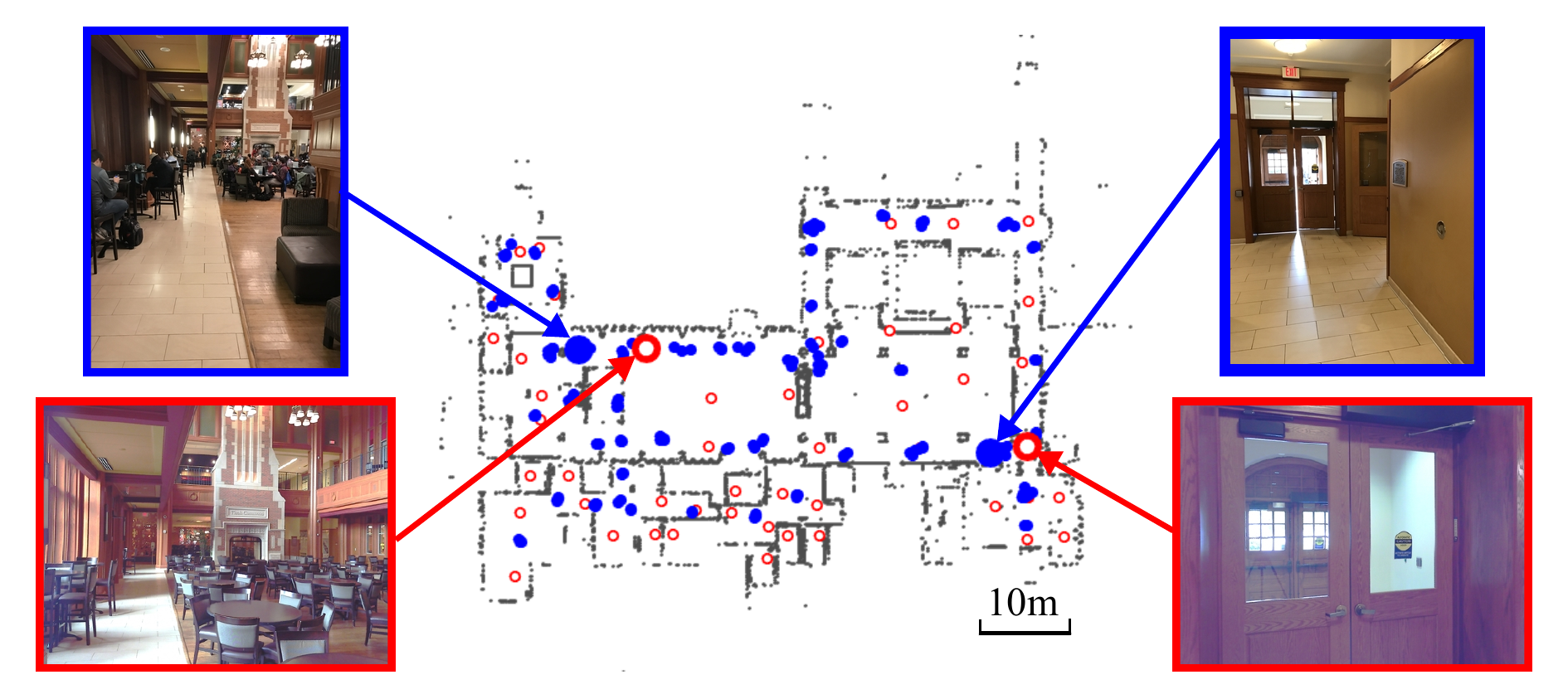} \\
    (a) DUC1 (first floor) \\[3pt]
    \includegraphics[width=0.9\linewidth]{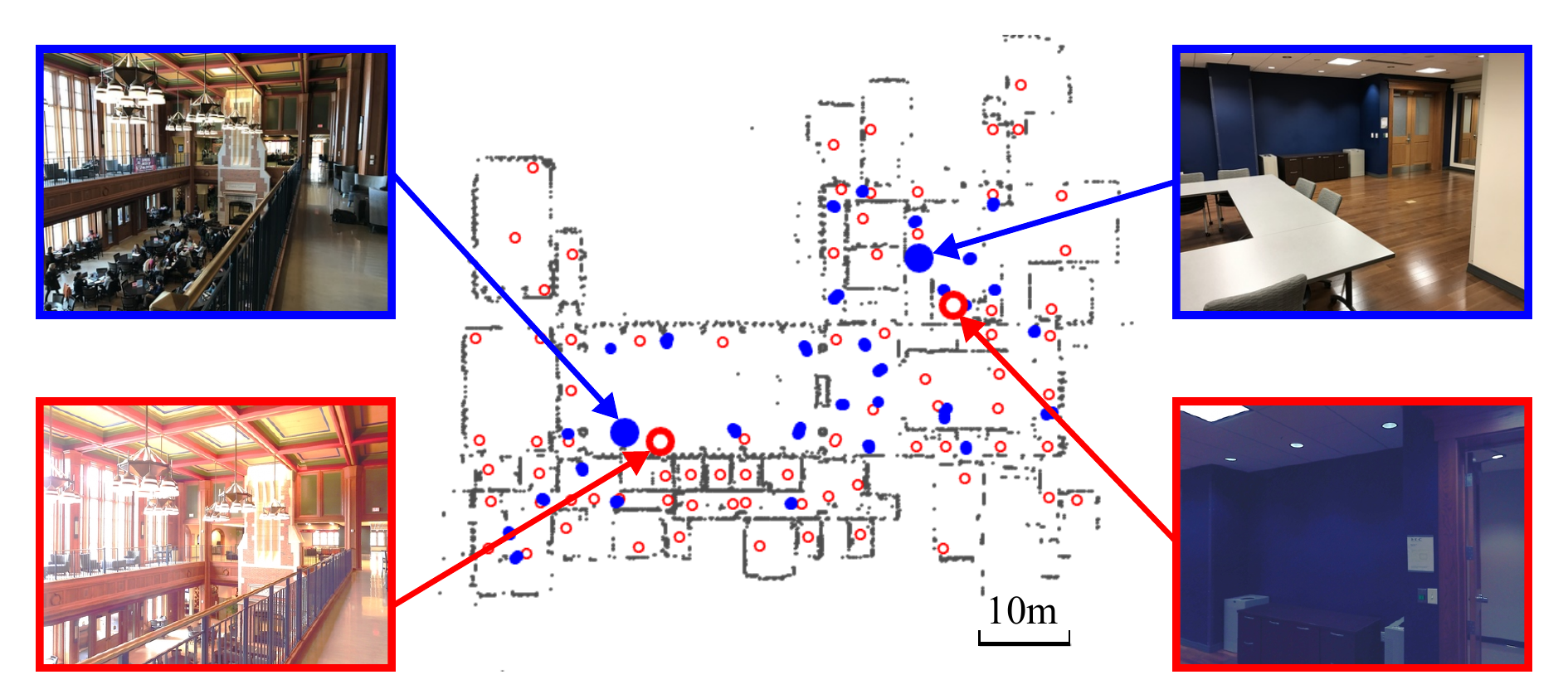} \\    
    (b) DUC2 (second floor) \\[3pt]
    \end{tabular}
    \caption{{\bf Query reference positions in the InLoc dataset. } The 329 reference poses of query images (blue dots) are plotted on the 3D maps (grey dots) that are generated by panoramic 3D scans at 277 distinct positions (red circles).
    }
    \label{fig:refposes_all}
\end{figure*}
}

{\tabcolsep=1pt
\begin{figure*}
\centering
\begin{tabular}{cccccc}
\includegraphics[width=0.16\linewidth]{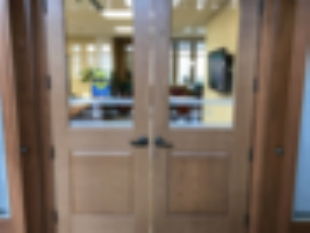} & 
\includegraphics[width=0.16\linewidth]{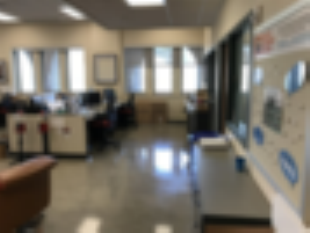} & 
\includegraphics[width=0.16\linewidth]{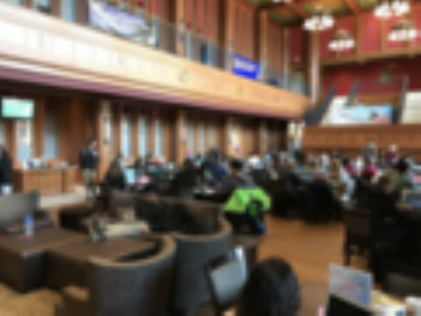} & 
\includegraphics[width=0.16\linewidth]{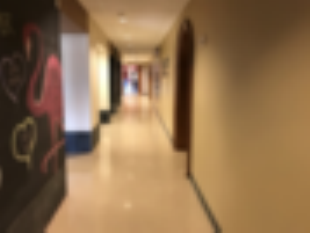} & 
\includegraphics[width=0.16\linewidth]{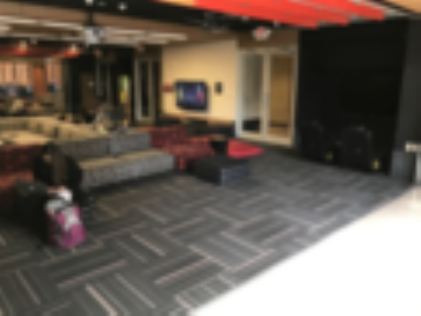} & 
\includegraphics[width=0.16\linewidth]{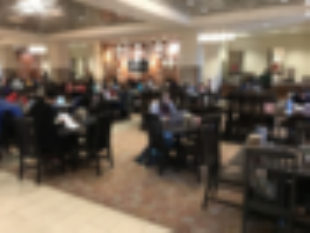} \\
\includegraphics[width=0.16\linewidth]{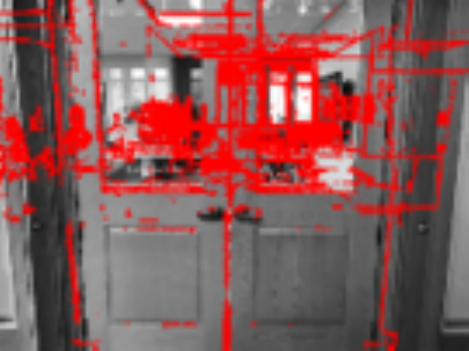} & 
\includegraphics[width=0.16\linewidth]{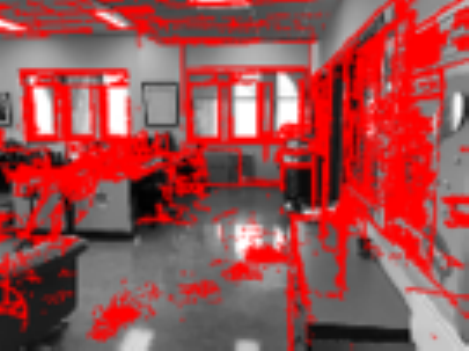} & 
\includegraphics[width=0.16\linewidth]{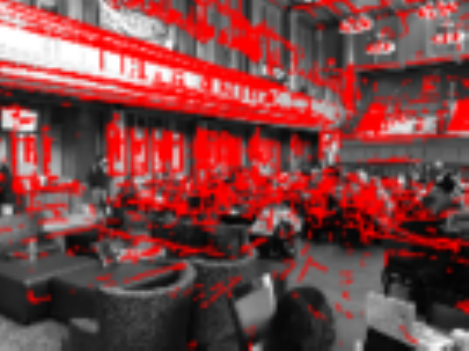} & 
\includegraphics[width=0.16\linewidth]{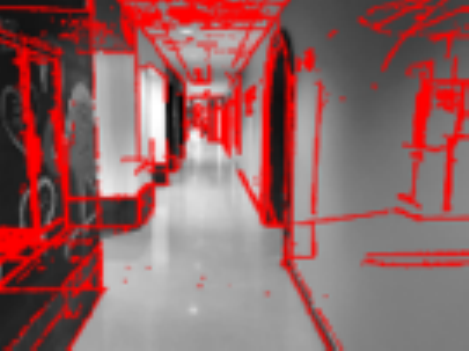} & 
\includegraphics[width=0.16\linewidth]{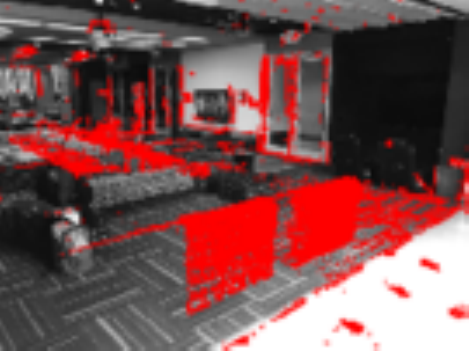} & 
\includegraphics[width=0.16\linewidth]{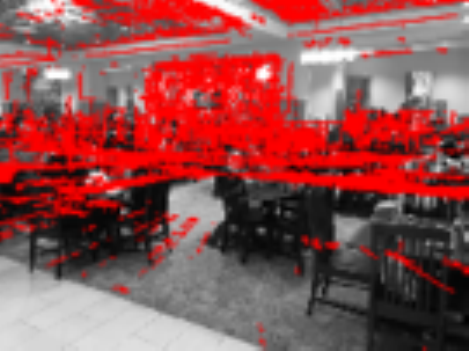} \\ [1pt] 
\hline \\ [-1.5ex]
\includegraphics[width=0.16\linewidth]{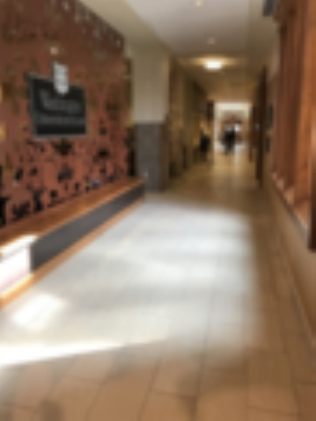} & 
\includegraphics[width=0.16\linewidth]{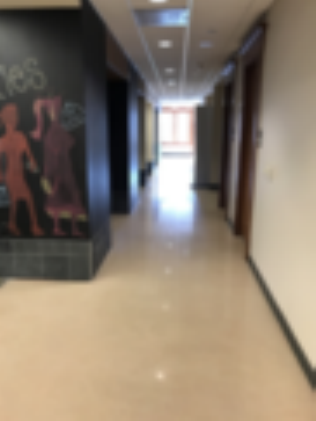} & 
\includegraphics[width=0.16\linewidth]{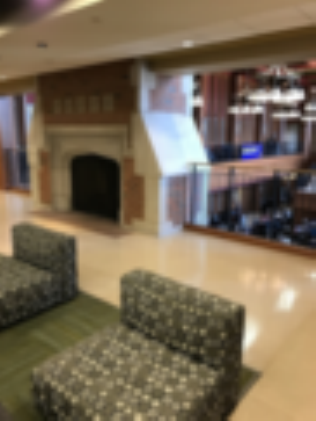} & 
\includegraphics[width=0.16\linewidth]{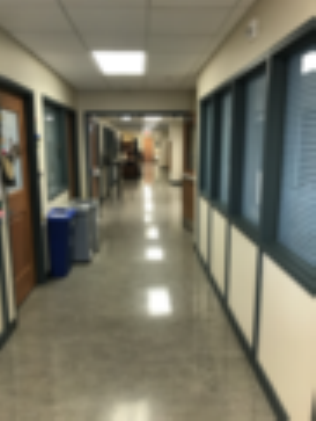} & 
\includegraphics[width=0.16\linewidth]{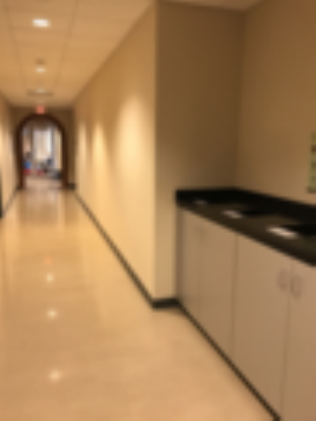} & 
\includegraphics[width=0.16\linewidth]{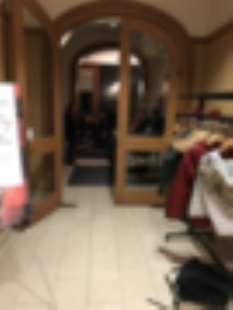} \\
\includegraphics[width=0.16\linewidth]{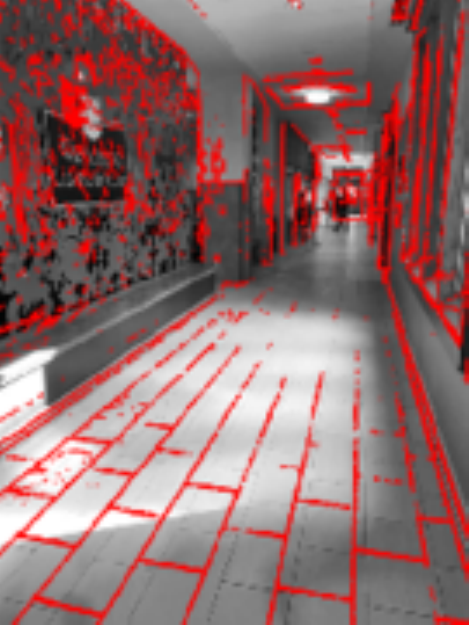} & 
\includegraphics[width=0.16\linewidth]{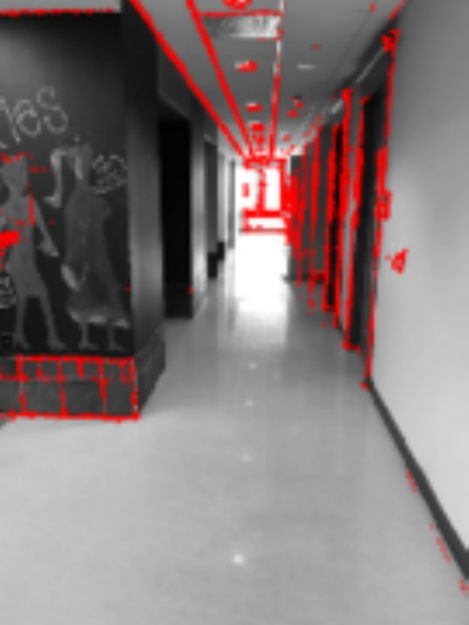} & 
\includegraphics[width=0.16\linewidth]{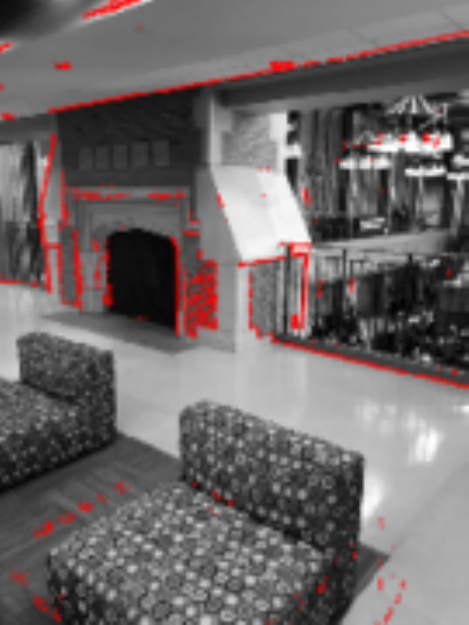} & 
\includegraphics[width=0.16\linewidth]{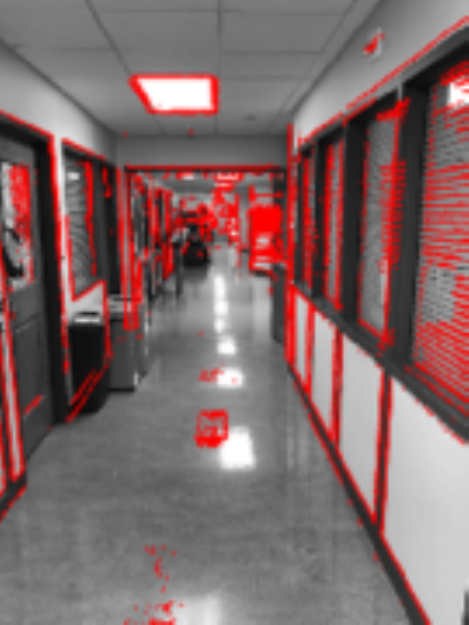} & 
\includegraphics[width=0.16\linewidth]{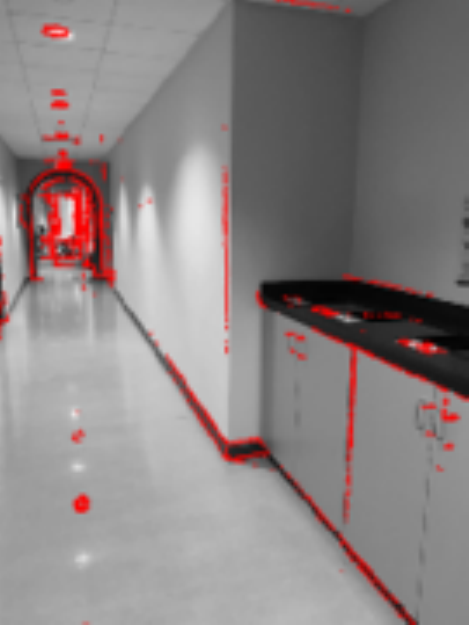} & 
\includegraphics[width=0.16\linewidth]{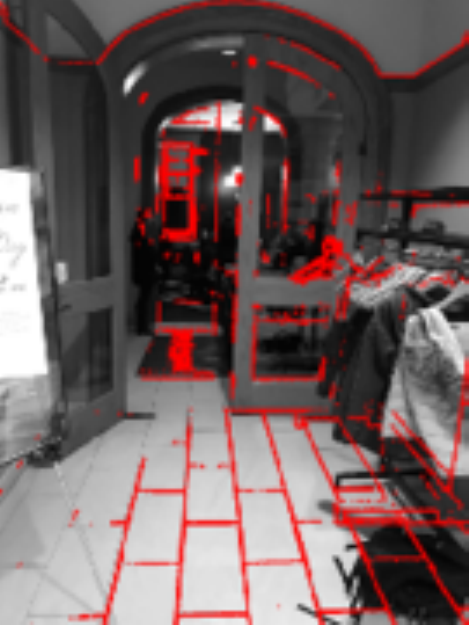} \\[2pt] 
\end{tabular}
\caption{{\bf Examples of query images and verified reference poses. } 
Each of the two groups show query images on top, followed by the same image, with database edges projected onto the queries. 
\label{fig:refposes_qual}}
\end{figure*}
}

\section{Qualitative results \label{sec:qual}}
\noindent 
In what follows, we will consider the query image correctly localized, if the error for the estimated pose is within 1 meter and 5$^\circ$ with respect to the reference pose. 

We first consider situations in which InLoc successfully localizes the query images, while the state-of-the-art NetVLAD+SparsePE fails. 
Figure~\ref{fig:qualitative_SPSvsPV} shows qualitative examples of the results obtained by NetVLAD+SparsePE (a,c,e) versus our InLoc (b,d,f). 
As shown in (a) and (c), sparse features are often detected on highly repetitive structures \eg, fonts (text), textured surfaces (fabric pattern on the sofa). As shown in (a) for the baseline, matching features found on such objects can result in matches with unrelated parts of the scene, leading to incorrect camera pose estimates. 
The fact that sparse features are predominantly found in few textured regions leads to problems in the largely untextured indoor scenes. 
This is shown in (e), where matches are found only in a small part of the query image, which leads to an unstable configuration for camera pose estimation. 
This, in turn, leads to more stable pose estimates in (b), (d), and (f). 
Our pose verification, {\bf DensePV} (section~\ref{sec:pose_verification}), allows us to identify incorrect poses, resulting from features found on repetitive structures, since most parts of the image rendered from a false pose are not consistent with the query image. Thus, InLoc is better suited to handle highly repetitive indoor scenes with rich feature correspondences. 

The next set of qualitative results demonstrates the benefits of dense pose verification. 
For this, figure~\ref{fig:qualitative_DNSvsPV} compares results obtained by InLoc (b,d,f) with results obtained by baseline NetVLAD+{\bf DensePE} (a,c,e). 
In this case, the baseline NetVLAD+{\bf DensePE} 
uses our dense matching ({\bf DensePE}) but selects the best pose based only on the number of inlier matches and not using our pose verification by virtual view synthesis ({\bf DensePV}). 
For scenes dominated by symmetries and repetitive structures (a,c) or largely texture-less regions (a, e), there can be a large amount of geometrically consistent matches even for unrelated database images. 
This still holds true 
even if matches are obtained by dense features and geometrically verified. 
Our dense pose verification strategy using synthesized images (b,d,f) effectively provides ``negative" evidence in such situations. The error maps (bottom row) clearly show that it detects (in)consistent areas between the query and its synthesized image. 

\para{Limitations. }
Our pose verification (section~\ref{sec:pose_verification}) evaluates the estimated camera pose by dense pixel-level matching between the query image and the synthesized view. 
This verification is robust up to a certain level of scene changes, \eg, illumination changes and some amount of misalignment, but cannot deal with extreme changes in the scene such as very large occlusions or when the view is dominated by moving objects. 

Figure~\ref{fig:qualitative_limitation} shows typical failure cases of InLoc, due to our pose verification not being able to identify the correct pose in highly dynamic scenes. 
In both cases, the query images capture many moving objects, \eg, people (a) or chairs (b), and highly dynamic scenes, \eg, opened/closed shutters (a) or pictures on the wall/removed (b). 
These moving objects cover a large part of the image.  

Those are the remaining open-issues that can be potentially addressed by adopting further semantic information~\cite{anand2013contextually,kim2017learned}. 

{\tabcolsep=1pt
\begin{figure*}
    \centering
    {\footnotesize
    \begin{tabular}{cc|cc|cc}
    \includegraphics[width=0.16\linewidth]{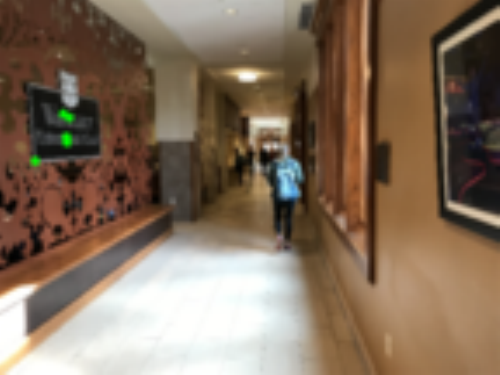} & 
    \includegraphics[width=0.16\linewidth]{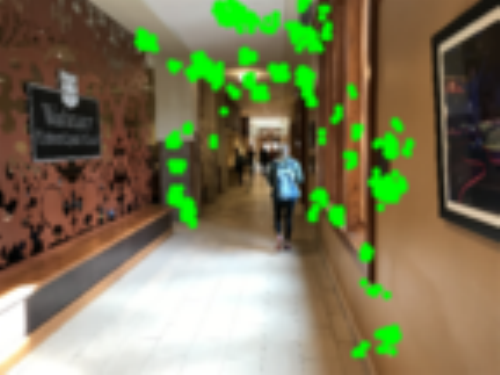} & 
    \includegraphics[width=0.16\linewidth]{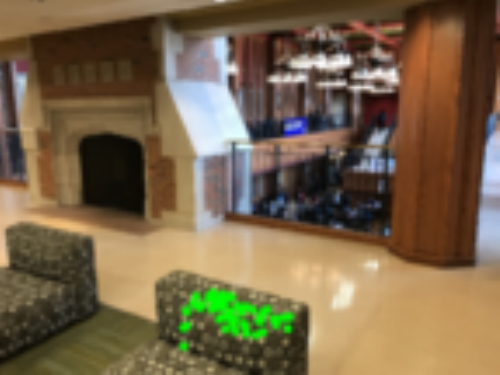} & 
    \includegraphics[width=0.16\linewidth]{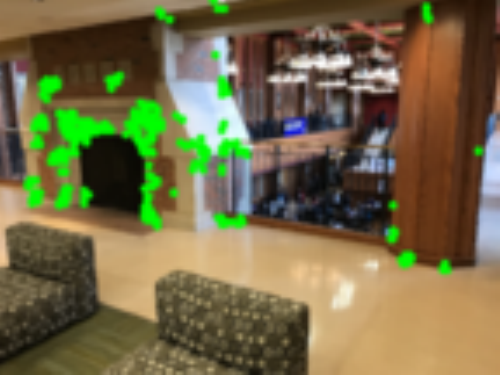} & 
    \includegraphics[width=0.16\linewidth]{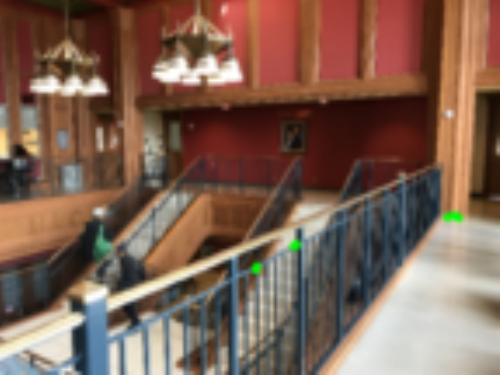} & 
    \includegraphics[width=0.16\linewidth]{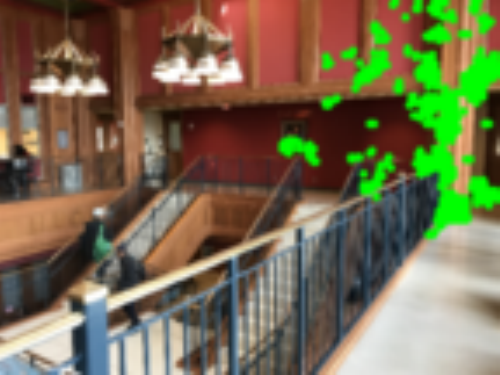} \\[-1pt]
    \includegraphics[width=0.16\linewidth]{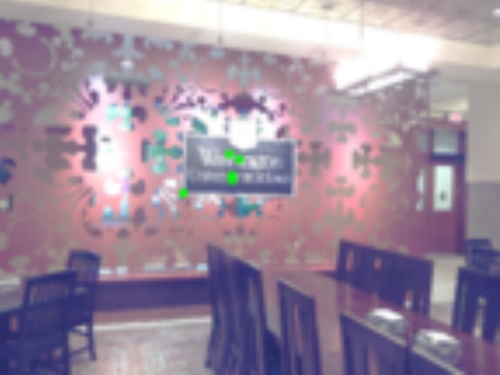} & 
    \includegraphics[width=0.16\linewidth]{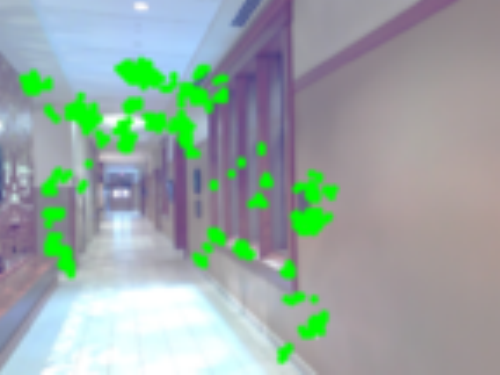} & 
    \includegraphics[width=0.16\linewidth]{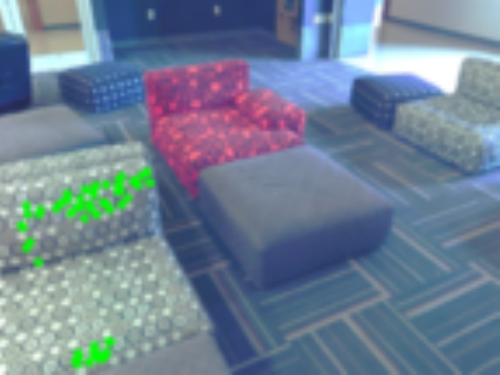} & 
    \includegraphics[width=0.16\linewidth]{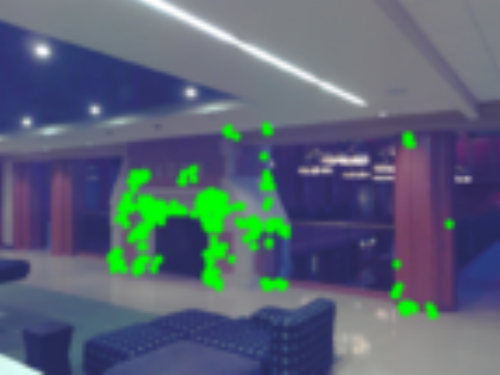} & 
    \includegraphics[width=0.16\linewidth]{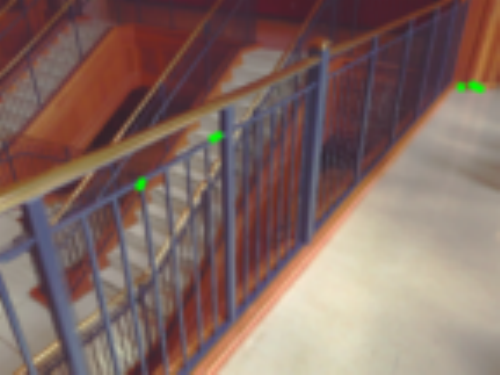} & 
    \includegraphics[width=0.16\linewidth]{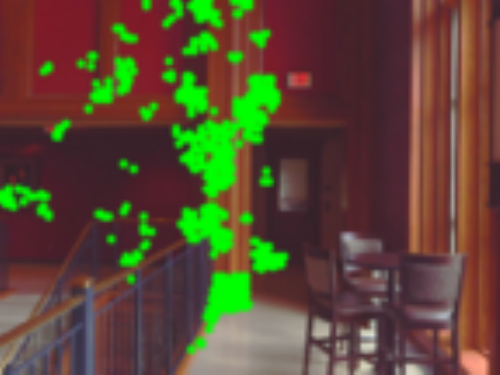} \\[-1pt]
    \includegraphics[width=0.16\linewidth]{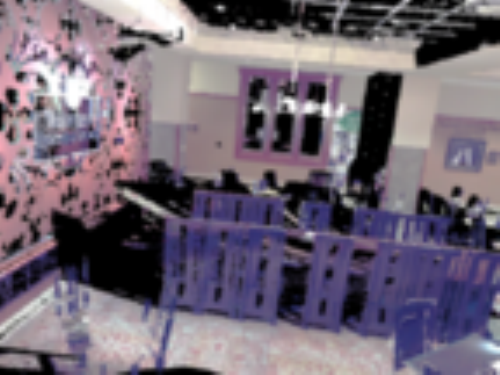} & 
    \includegraphics[width=0.16\linewidth]{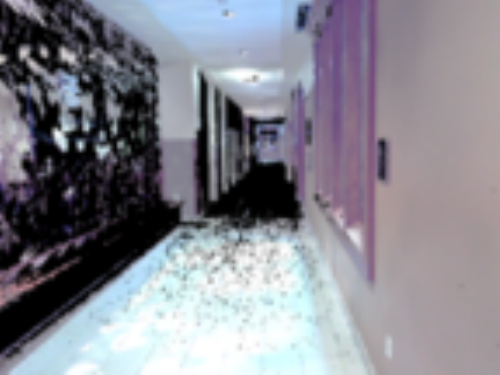} & 
    \includegraphics[width=0.16\linewidth]{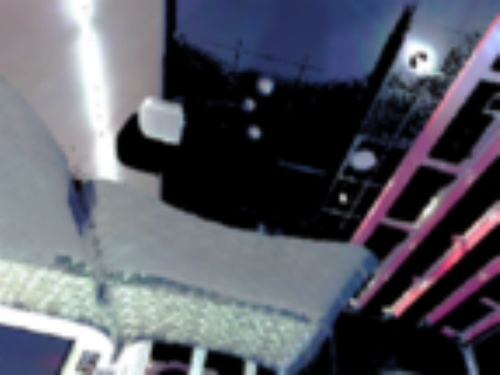} & 
    \includegraphics[width=0.16\linewidth]{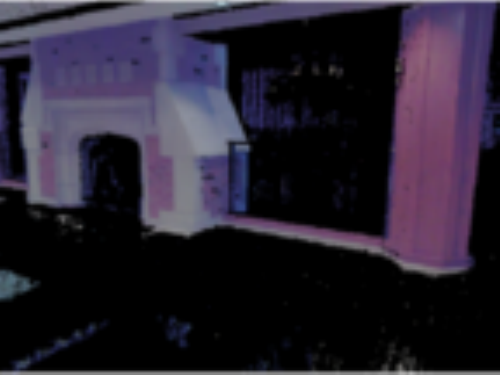} & 
    \includegraphics[width=0.16\linewidth]{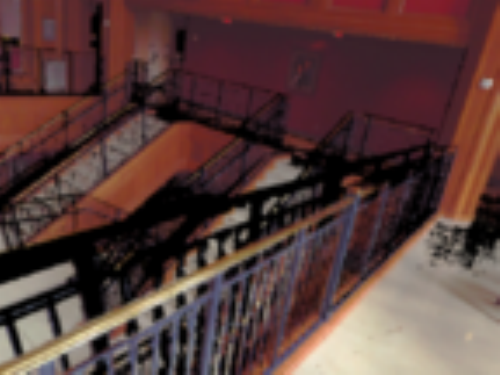} & 
    \includegraphics[width=0.16\linewidth]{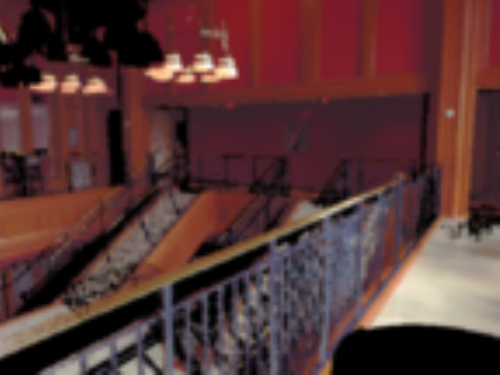} \\[-1pt]
    \includegraphics[width=0.16\linewidth]{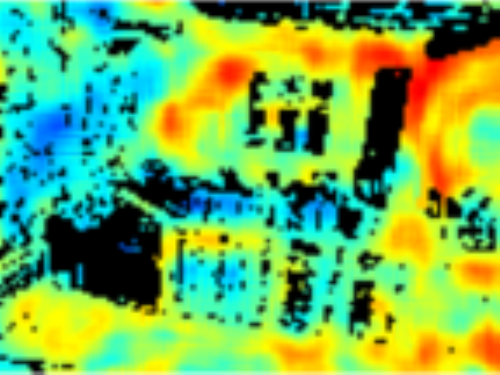} & 
    \includegraphics[width=0.16\linewidth]{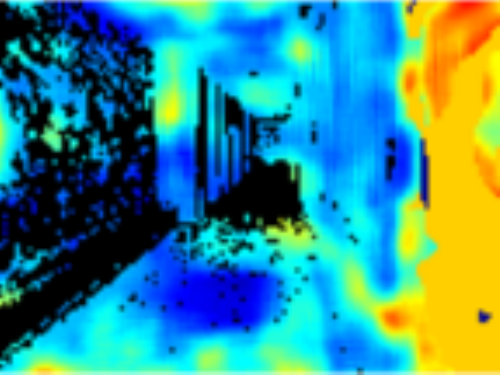} & 
    \includegraphics[width=0.16\linewidth]{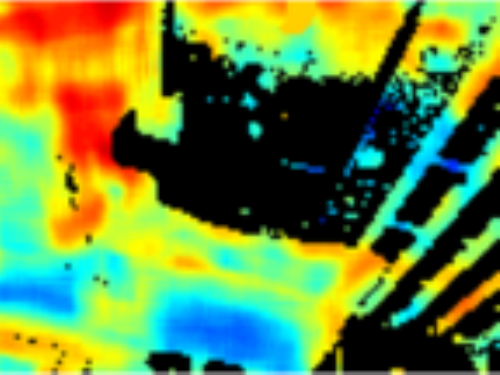} & 
    \includegraphics[width=0.16\linewidth]{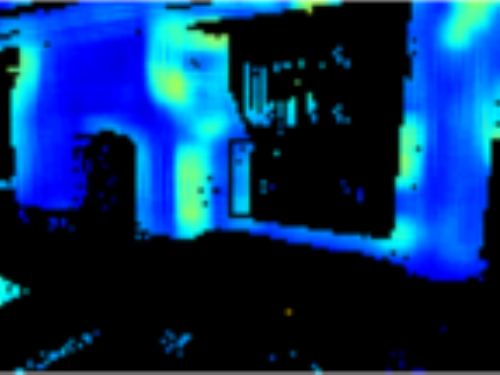} & 
    \includegraphics[width=0.16\linewidth]{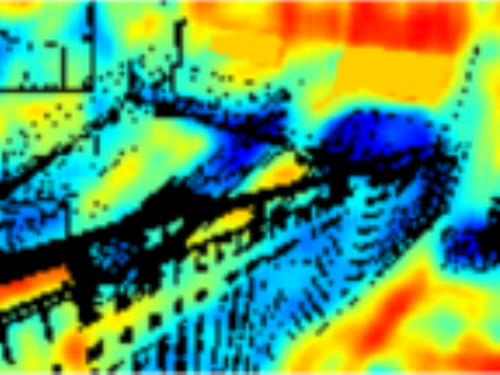} & 
    \includegraphics[width=0.16\linewidth]{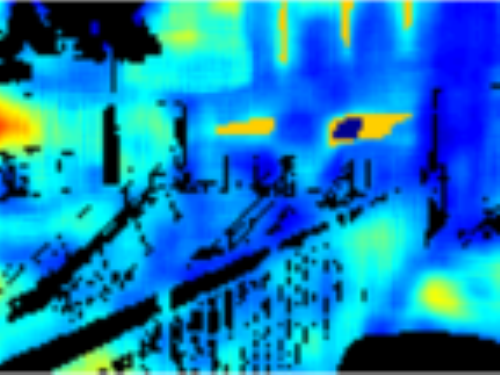} \\[-1pt]
    10.51, 178.81$^{\circ}$ & 0.06, 1.00$^{\circ}$ & 16.59, 133.63$^{\circ}$ & 0.19, 2.46$^{\circ}$ & 2.27, 17.39$^{\circ}$ & 0.46, 2.89$^{\circ}$ \\[1pt] \hline
    NetVLAD~\cite{Arandjelovic16}+ & 
    {\bf InLoc:} NetVLAD~\cite{Arandjelovic16}+ & 
    NetVLAD~\cite{Arandjelovic16}+ & 
    {\bf InLoc:} NetVLAD~\cite{Arandjelovic16}+ & 
    NetVLAD~\cite{Arandjelovic16}+ & 
    {\bf InLoc:} NetVLAD~\cite{Arandjelovic16}+ \\
    SparsePE & 
    {\bf DensePE}+{\bf DensePV} & 
    SparsePE & 
    {\bf DensePE}+{\bf DensePV} & 
    SparsePE & 
    {\bf DensePE}+{\bf DensePV}\\[2pt]  
    \normalsize{(a)} & \normalsize{(b)} & 
    \normalsize{(c)} & \normalsize{(d)} & 
    \normalsize{(e)} & \normalsize{(f)} \\[2pt]
    \end{tabular}
    \caption{
    {\bf Qualitative comparison between InLoc and NetVLAD+SparsePE. } In these examples, InLoc successfully localizes a query image within 1 meter distance error and 5$^{\circ}$ angular error with respect to the reference pose whereas the state-of-the-art NetVLAD+SparsePE fails. From top to bottom: query image, the best matching database image, synthesized view rendered from the estimated pose, error map between the query image and the synthesized view, localization error (meters, degrees). Warm colors correspond to large errors. Green dots are the inlier matches obtained by P3P-LO-RANSAC.  \label{fig:qualitative_SPSvsPV}}
    }
\end{figure*}
}

{\tabcolsep=1pt
\begin{figure*}
    \centering
    {\footnotesize
    \begin{tabular}{cc|cc|cc}
    \includegraphics[width=0.16\linewidth]{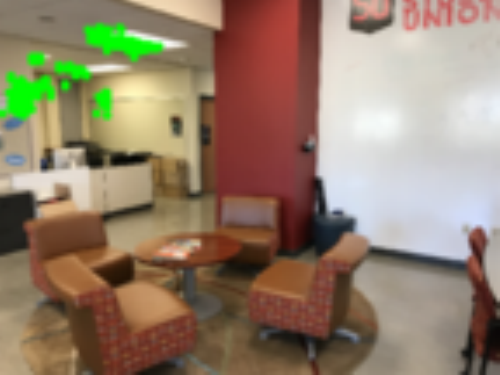} & 
    \includegraphics[width=0.16\linewidth]{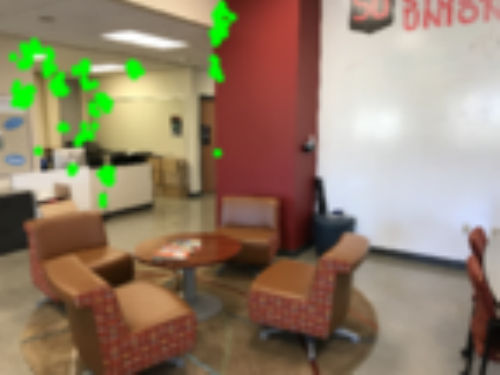} & 
    \includegraphics[width=0.16\linewidth]{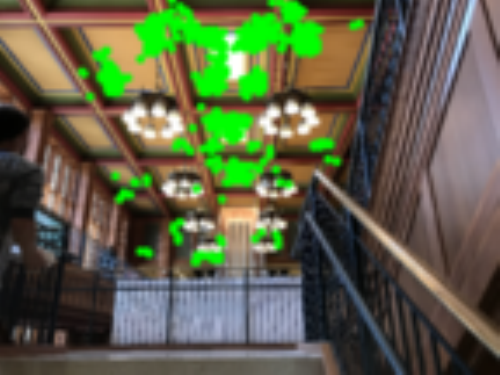} & 
    \includegraphics[width=0.16\linewidth]{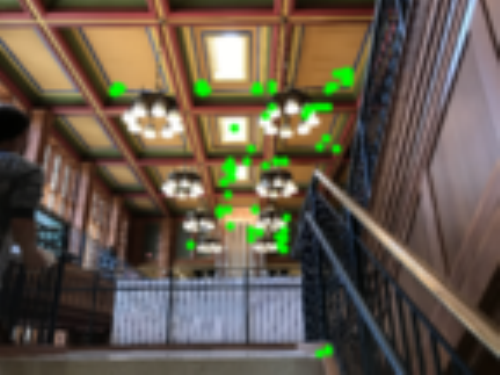} & 
    \includegraphics[width=0.16\linewidth]{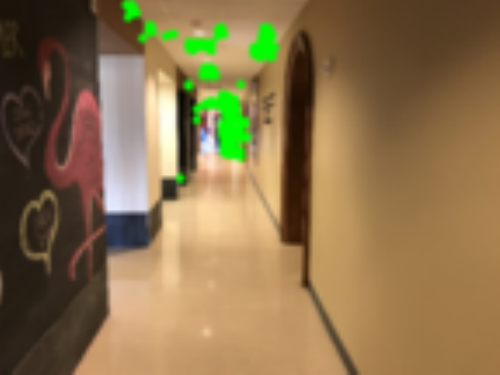} & 
    \includegraphics[width=0.16\linewidth]{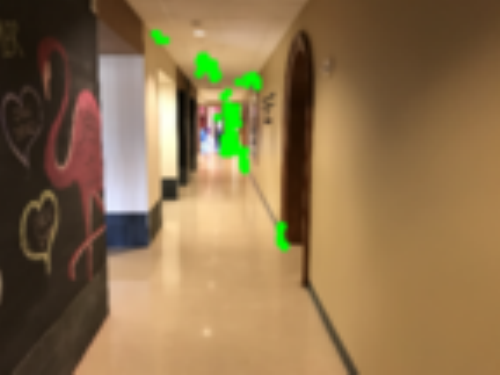} \\[-1pt]
    \includegraphics[width=0.16\linewidth]{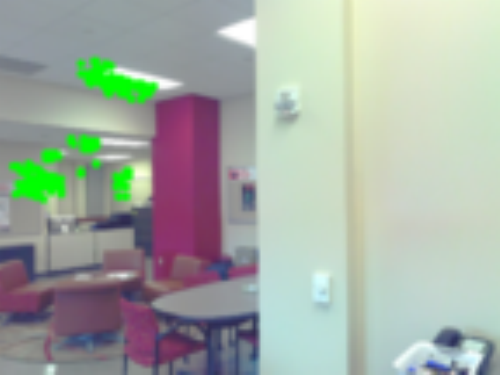} & 
    \includegraphics[width=0.16\linewidth]{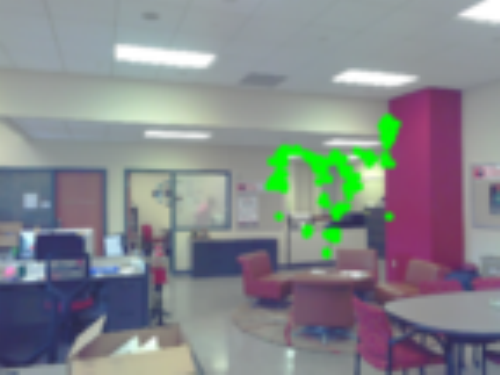} & 
    \includegraphics[width=0.16\linewidth]{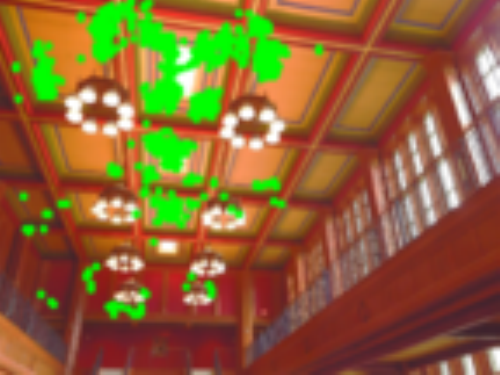} & 
    \includegraphics[width=0.16\linewidth]{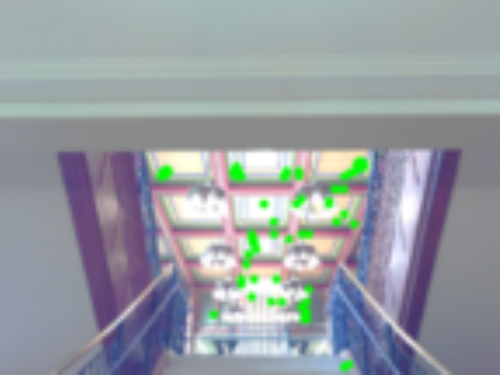} & 
    \includegraphics[width=0.16\linewidth]{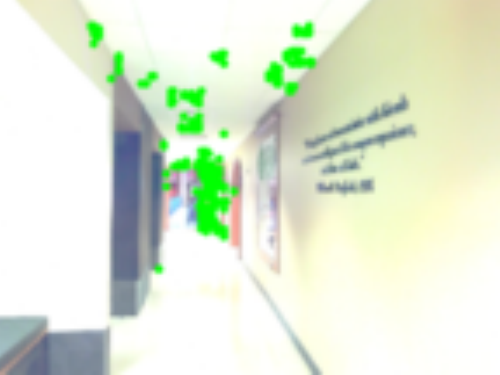} & 
    \includegraphics[width=0.16\linewidth]{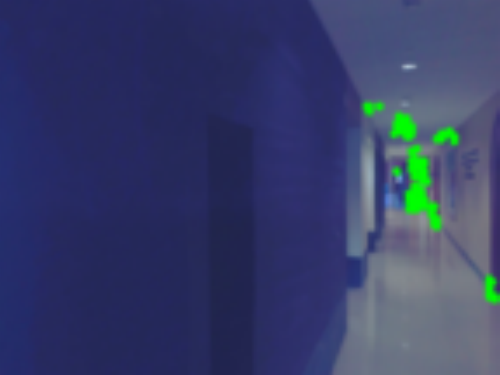} 
    \\[-1pt]
    \includegraphics[width=0.16\linewidth]{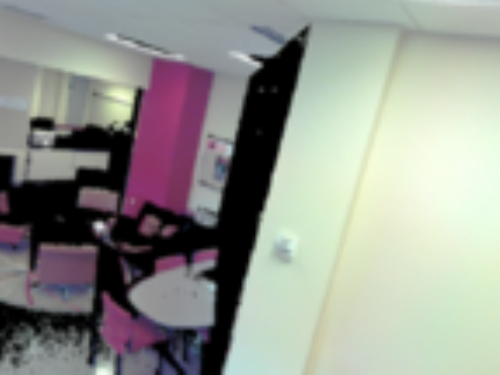} & 
    \includegraphics[width=0.16\linewidth]{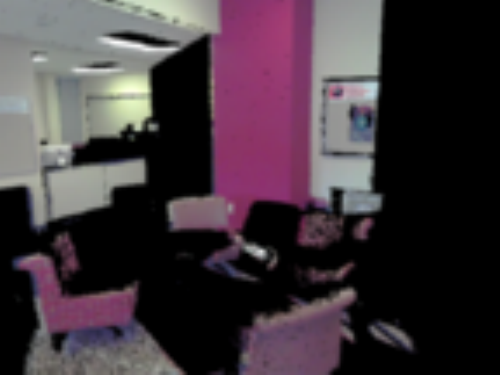} & 
    \includegraphics[width=0.16\linewidth]{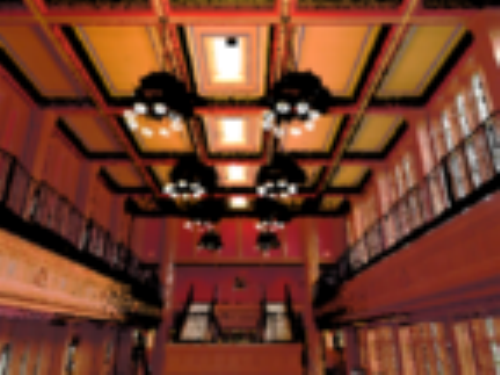} & 
    \includegraphics[width=0.16\linewidth]{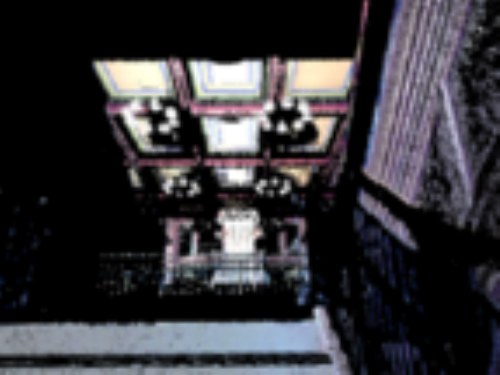} & 
    \includegraphics[width=0.16\linewidth]{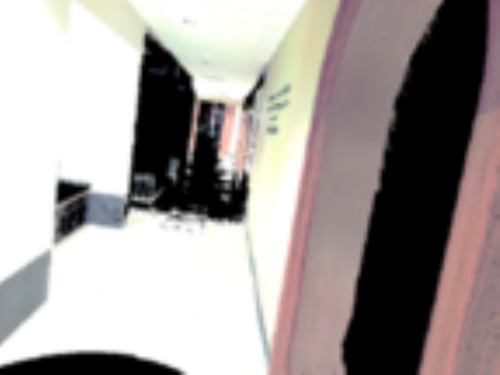} & 
    \includegraphics[width=0.16\linewidth]{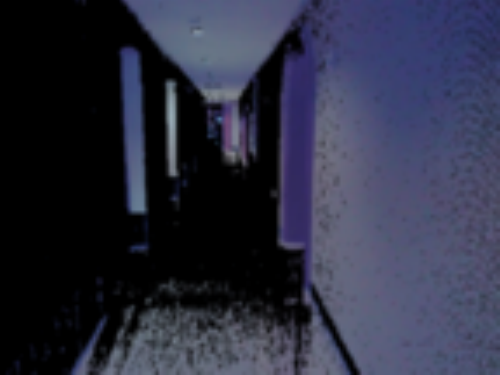} \\[-1pt]
    \includegraphics[width=0.16\linewidth]{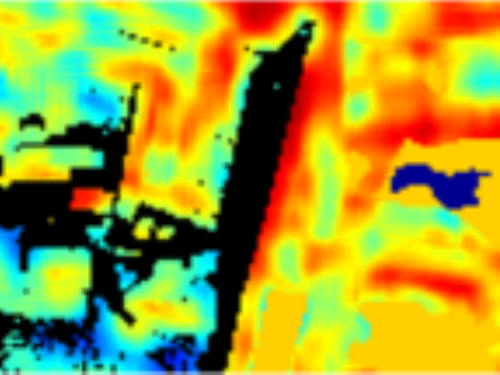} & 
    \includegraphics[width=0.16\linewidth]{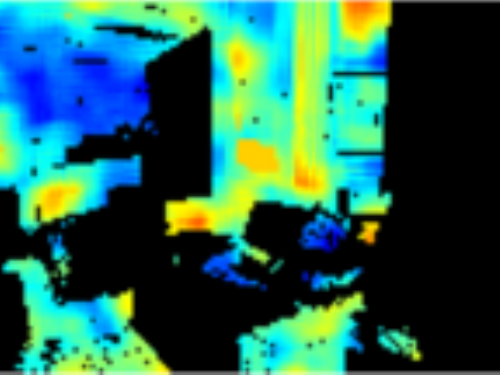} & 
    \includegraphics[width=0.16\linewidth]{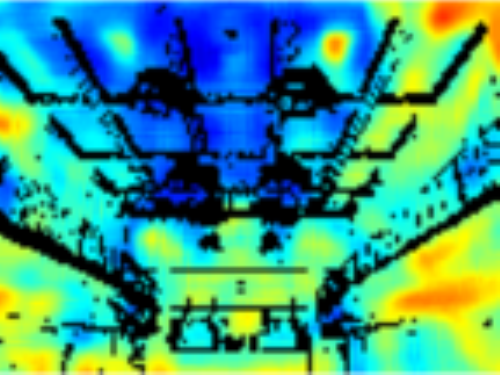} & 
    \includegraphics[width=0.16\linewidth]{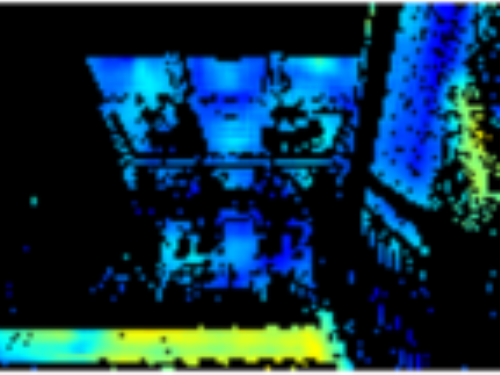} & 
    \includegraphics[width=0.16\linewidth]{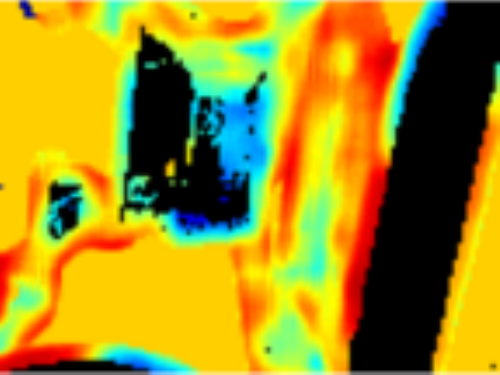} & 
    \includegraphics[width=0.16\linewidth]{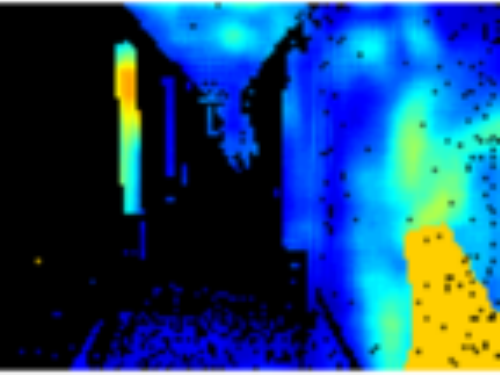} \\[-1pt]
    3.74, 14.56$^{\circ}$ & 0.21, 1.25$^{\circ}$ & 24.51, 179.61$^{\circ}$ & 0.49, 1.50$^{\circ}$ & 3.15, 8.17$^{\circ}$ & 0.20, 1.19$^{\circ}$ \\[1pt] \hline
    NetVLAD~\cite{Arandjelovic16}+ & 
    {\bf InLoc:} NetVLAD~\cite{Arandjelovic16}+ & 
    NetVLAD~\cite{Arandjelovic16}+ & 
    {\bf InLoc:} NetVLAD~\cite{Arandjelovic16}+ & 
    NetVLAD~\cite{Arandjelovic16}+ & 
    {\bf InLoc:} NetVLAD~\cite{Arandjelovic16}+ \\
    {\bf DensePE} & 
    {\bf DensePE}+{\bf DensePV} & 
    {\bf DensePE} & 
    {\bf DensePE}+{\bf DensePV} & 
    {\bf DensePE} & 
    {\bf DensePE}+{\bf DensePV}\\[2pt]
    \normalsize{(a)} & \normalsize{(b)} & 
    \normalsize{(c)} & \normalsize{(d)} & 
    \normalsize{(e)} & \normalsize{(f)} \\[2pt]
    \end{tabular}
    \caption{
    {\bf Qualitative comparison between InLoc and NetVLAD+DensePE. } InLoc successfully localizes a query image within 1 meter distance error and 5$^{\circ}$ angular error with respect to the reference pose whereas NetVLAD+DensePE (no pose verification via synthesis), fails. From top to bottom: query image, the best matching database image, synthesized view rendered from the estimated pose, error map between the query image and the synthesized view, localization error (meters, degrees). Warm colors correspond to large errors. Green dots are the inlier matches obtained by P3P-LO-RANSAC.  \label{fig:qualitative_DNSvsPV}}
    }
\end{figure*}
}

{\tabcolsep=1pt
\begin{figure}
    \centering
    {\footnotesize
    \begin{tabular}{c@{\hskip 4pt}c}
    \includegraphics[width=0.48\linewidth]{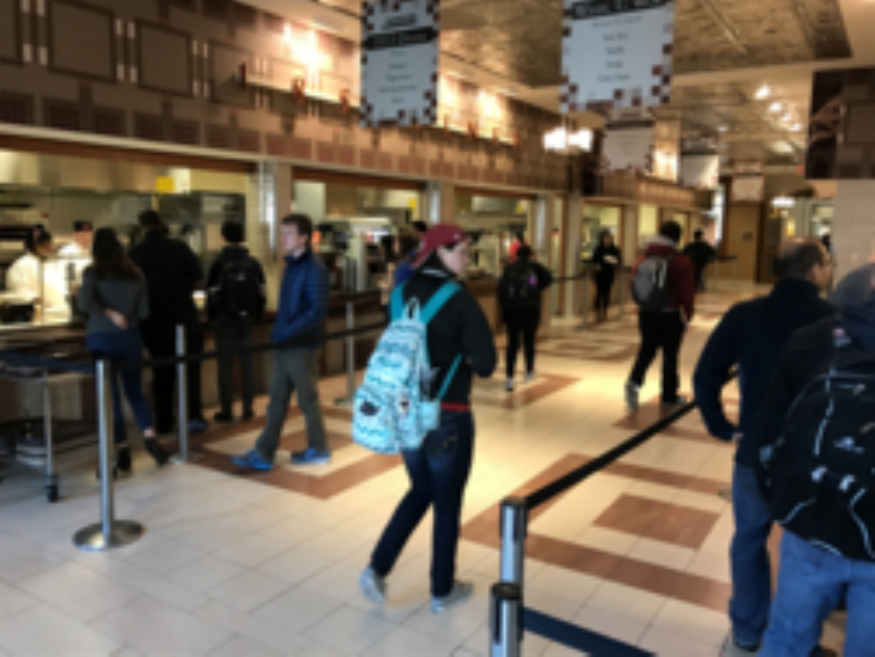} & 
    \includegraphics[width=0.32\linewidth]{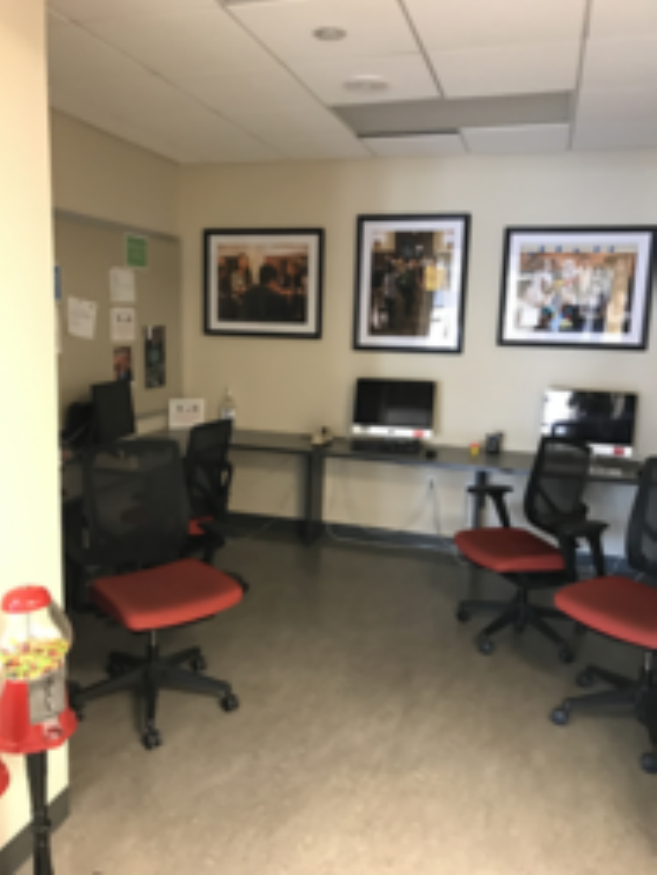}  
    \\[2pt]
    \includegraphics[width=0.48\linewidth]{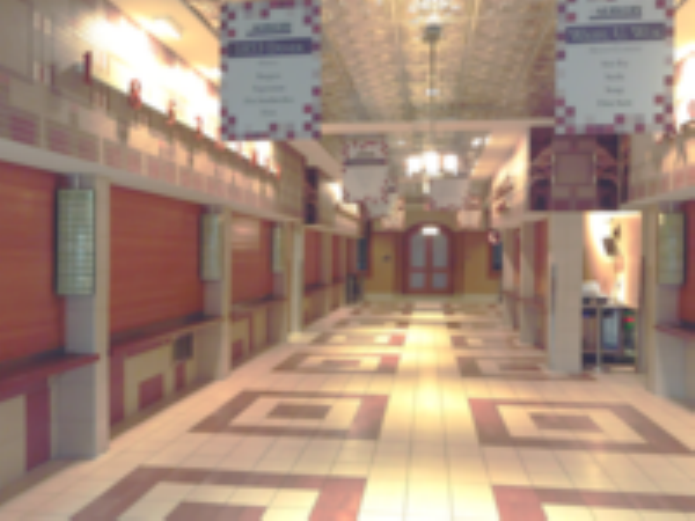} & 
    \includegraphics[width=0.48\linewidth]{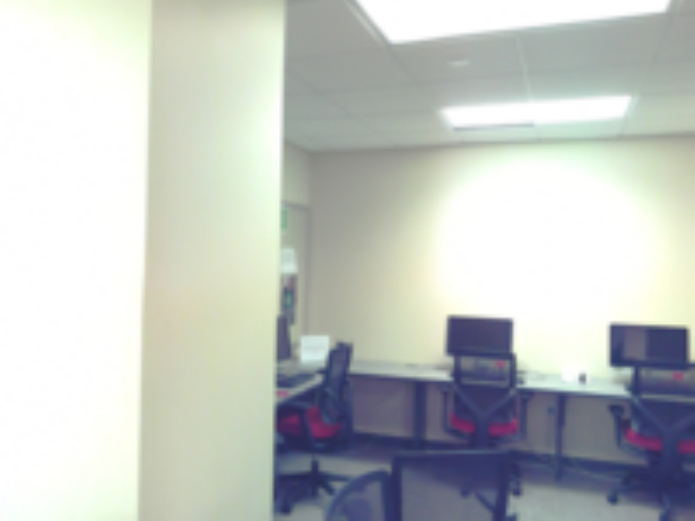}  
    \\[2pt]
    \normalsize{(a)} & \normalsize{(b)}
    \\[2pt]
    \end{tabular}
    \caption{
    {\bf Failure cases. } Our InLoc approach fails to localize these examples due to many moving objects, \eg people (a) or chairs (b), and highly dynamic scenes, \eg opened/closed shutters (a) or pictures on the wall/removed (b). From top to bottom: query image and the reference database image.
    \label{fig:qualitative_limitation}}
    }
\end{figure}
}

{\small
\bibliographystyle{ieee}
\bibliography{shortstrings,vgg_local,vgg_other,taira}
}

\end{document}